\pdfoutput=1

\documentclass[a4paper]{cas-sc}

\usepackage{hyperref}
\usepackage{xcolor}
\hypersetup{
	colorlinks=true,
	linkcolor=blue,
	filecolor=blue,
	urlcolor=black,
	citecolor=blue
}
\usepackage[all]{hypcap}
\usepackage[numbers]{natbib}
\usepackage{amsmath}
\usepackage{amssymb}
\usepackage{graphicx}
\usepackage{booktabs}   
\usepackage{float}	
\usepackage{times}
\usepackage{epsfig}
\usepackage{subfigure}
\usepackage{graphicx}
\usepackage{subcaption}
\usepackage{float}
\usepackage{overpic}
\usepackage{ragged2e}
\usepackage{enumitem}
\usepackage{multirow}
\usepackage{caption}
\usepackage{natbib}
\usepackage{xcolor}
\usepackage{soul}
\usepackage{amsmath}
\usepackage{mathtools}
\definecolor{revisioncolor}{HTML}{FFE2E2}

\def\tsc#1{\csdef{#1}{\textsc{\lowercase{#1}}\xspace}}
\tsc{WGM}
\tsc{QE}

\begin{document}

\setcitestyle{sort&compress}

\let\WriteBookmarks\relax
\def\floatpagepagefraction{1}
\def\textpagefraction{.001}
\let\printorcid\relax

\shorttitle{<short title of the paper for running head>} 
\shorttitle{Ambiguity-Aware and High-Order Relation Learning for Multi-Grained Image-Text Matching}    

\shortauthors{<short author list for running head>}
\shortauthors{Junyu Chen et al.}

\title[mode = title]{Ambiguity-Aware and High-Order Relation Learning for Multi-Grained Image-Text Matching} 

\author[1]{Junyu Chen}
\ead{2023210516022@stu.cqnu.edu.cn}

\author[1]{Yihua Gao}
\ead{2023210516033@stu.cqnu.edu.cn}

\author[2]{Mingyuan Ge}
\ead{20242401025@stu.cqu.edu.cn}

\author[1]{Mingyong Li}
\ead{limingyong@cqnu.edu.cn}
\cormark[1]

\address[1]{College of Computer and Information Science, Chongqing Normal University, Chongqing 401331, China}
\address[2]{School of Big Data and Software Engineering, Chongqing University, Chongqing 401331, China}

\cortext[1]{Corresponding author}

\begin{abstract}
	Image-text matching is crucial for bridging the semantic gap between computer vision and natural language processing. However, existing methods still face challenges in handling high-order associations and semantic ambiguities among similar instances. These ambiguities arise from subtle differences between soft positive samples (semantically similar but incorrectly labeled) and soft negative samples (locally matched but globally inconsistent), creating matching uncertainties.  Furthermore, current methods fail to fully utilize the neighborhood relationships among semantically similar instances within training batches, limiting the model's ability to learn high-order shared knowledge. This paper proposes the Ambiguity-Aware and High-order Relation learning framework (AAHR) to address these issues. AAHR constructs a unified representation space through dynamic clustering prototype contrastive learning, effectively mitigating the soft positive sample problem. The framework introduces global and local feature extraction mechanisms and an adaptive aggregation network, significantly enhancing full-grained semantic understanding capabilities. Additionally, AAHR employs intra-modal and inter-modal correlation matrices to investigate neighborhood relationships among sample instances thoroughly. It incorporates GNN to enhance semantic interactions between instances. Furthermore, AAHR integrates momentum contrastive learning to expand the negative sample set. These combined strategies significantly improve the model's ability to discriminate between features. Experimental results demonstrate that AAHR outperforms existing state-of-the-art methods on Flickr30K, MSCOCO, and ECCV Caption datasets, considerably improving the accuracy and efficiency of image-text matching. The code and model checkpoints for this research are available at \href{https://github.com/Image-Text-Matching/AAHR}{https://github.com/Image-Text-Matching/AAHR}.
\end{abstract}

\begin{highlights}
	\item Clustering prototype learning mitigates soft positive sample issues in image-text matching.
	\item Adaptive fusion of global-local features enables comprehensive semantic understanding.
	\item Graph-based modeling captures high-order relations among similar instances.
	\item Memory-augmented contrastive learning enhances cross-modal feature discrimination.
	\item State-of-the-art results on Flickr30K, MSCOCO and ECCV Caption benchmarks.
\end{highlights}


\maketitle

\section{Introduction}\label{sec:Introduction}
	\begin{figure}[t]
		\centering
		\includegraphics[width=\linewidth]{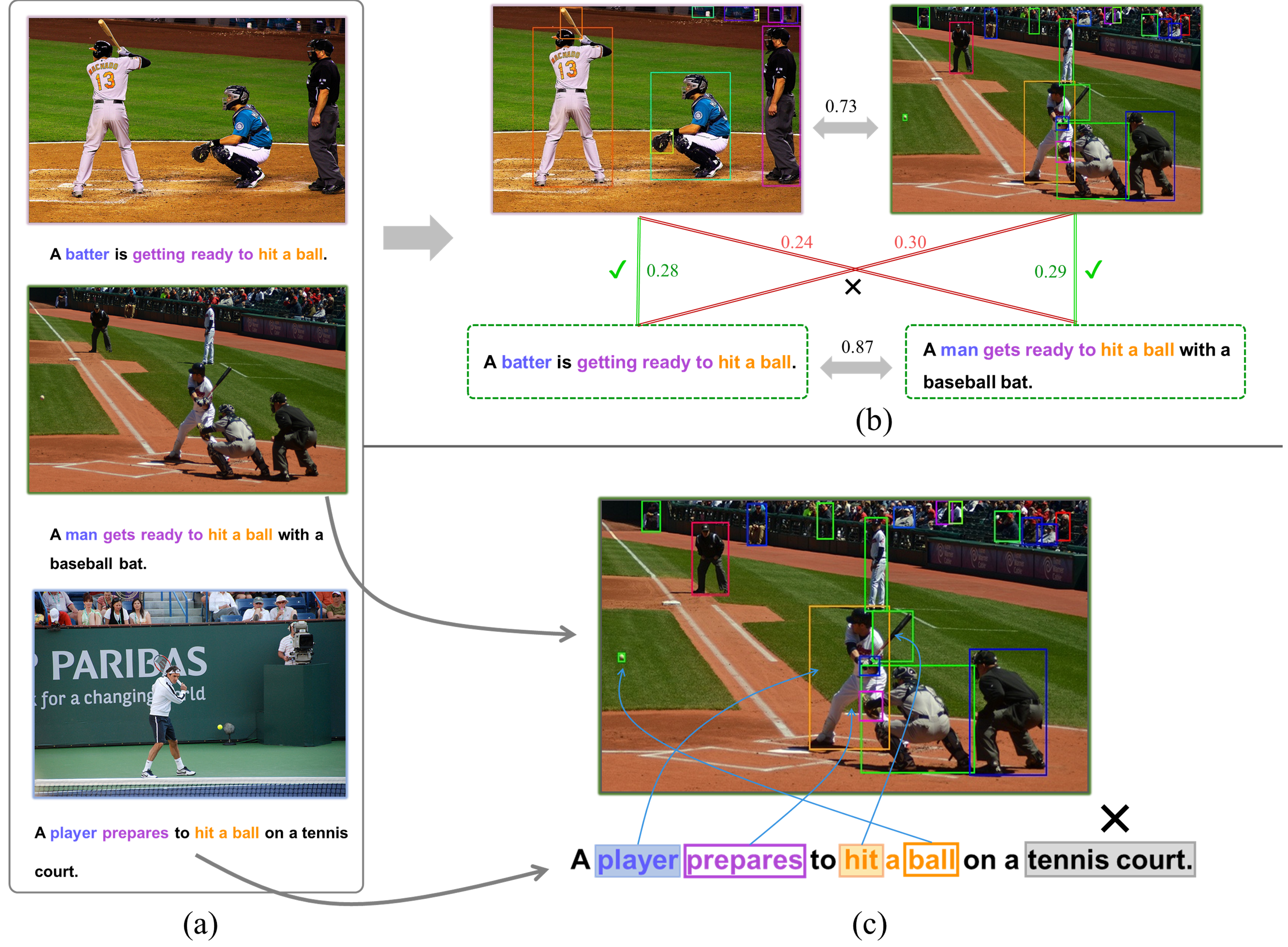}
		\caption{Illustrating our motivation. Exploring higher-order associations and subtle differences between semantically similar instances is beneficial for image-text matching tasks: (a) Intra-batch semantically similar instances: Despite differing in details (\textit{e.g.}, baseball vs. tennis), the core semantic concept of “preparing to hit the ball” is identical, containing rich higher-order shared knowledge. (b) Soft positive sample issue: Two image-text pairs describing similar baseball scenarios. Although semantically close, they are incorrectly labeled as negative samples. The numbers indicate the cosine similarity of embeddings obtained using our method. (c) Soft negative sample issue: One of the top-5 retrieval results using the CHAN method. The text matches well with the local details of the image, but the overall semantics are inconsistent.
		} \label{fig:motivation}
	\end{figure}

Image-text matching aims to achieve precise semantic alignment between image and text, laying a crucial foundation for multimodal learning. As a core technology in cross-modal understanding and generation, image-text matching plays a pivotal role in various applications, including cross-modal retrieval \cite{COTS, TMASS}, image captioning \cite{caption, p_caption}, visual grounding \cite{Clip_vg, HiVG}, and visual question answering \cite{Tem-adapter, KBQA}. Moreover, it serves as an essential supporting component for cutting-edge technologies such as modern multimodal large language models (MLLMs) \cite{LLaVA, Qwen2.5-VL} and text-to-image diffusion models \cite{LDM, synthesis}.
Existing methods can be primarily categorized into two types: global matching methods based on independent embeddings \cite{IMEB, VSE++, DSRAN, GPO, HREM, ESA} and local matching methods based on interactive embeddings \cite{SCAN, GSMN, SGRAF, NAAF, CHAN, BOOM}. The former encodes images and text into a shared embedding space, quantifying inter-modal semantic similarity by comparing global embeddings. This approach allows for the pre-computation of representations and caching indices, gaining popularity in recent large-scale retrieval applications due to its efficiency and considerable accuracy. The latter focuses on learning fine-grained alignments between salient visual elements in images and text fragments, capable of capturing more intricate semantic associations. However, these highly complex methods may face significantly increased inference time challenges.

Existing methods often overlook exploring high-order associations and subtle differences among semantically similar instances. Three main issues persist:

\begin{itemize}
	\item
	“Soft positive sample”: This issue is prevalent in the construction of current vision-language datasets (such as MSCOCO \cite{COCO}, Flickr30K \cite{f30k}, and LAION-400M \cite{LAION}), where semantically similar samples may be incorrectly labeled as negative, leading to confusion during model training and affecting evaluation accuracy, as shown in Fig. \ref{fig:motivation}(b). Addressing this issue, Chun et al. \cite{ECCV_Caption} pointed out that numerous soft positive samples compromise the accuracy of model performance evaluation. They developed the Extended COCO Validation (ECCV) Caption dataset through a combination of machine and human annotation, providing crucial support for the comprehensive assessment of model retrieval performance. However, constructing the ECCV Caption dataset requires substantial human and computational resources, and the correction was only applied to the MSCOCO test set. For larger training sets, completely rectifying such annotation bias remains impractical.
	\item
	“Soft negative sample”: Traditional local matching methods often suffer from limited global semantic awareness, overlooking crucial contextual information such as background and surrounding context \cite{DSRAN}. During retrieval, these methods frequently identify instances that exhibit strong local feature matches with the anchor but diverge significantly in global semantic meaning. As illustrated in Fig. \ref{fig:motivation}(c), CHAN \cite{CHAN}, the state-of-the-art open-source local matching method, produces top-5 retrieval results containing examples that, despite showing high similarity in local visual details, demonstrate notable discrepancies in global semantic expression.
	\item
	Neighborhood relation modeling: How to effectively model neighborhood relationships among semantically similar instances within training batches, enabling the model to learn high-order shared knowledge, remains an urgent problem to solve. As shown in Fig. \ref{fig:motivation}(a), the concept of “preparing to hit a ball” in different sports scenarios shares essential similarities despite potential variations in specific details. If a model could capture and generalize this high-order shared knowledge, it would significantly enhance its cross-scenario understanding and generalization capabilities.
\end{itemize}

We propose the Ambiguity-Aware and High-order Relation learning framework (AAHR) to address these issues. This framework first introduces a clustering prototype contrastive learning method \cite{CC, SWAV, DeepCluster}, dynamically learning unified prototypes in the joint image-text representation space through online clustering. These unified prototypes serve as bridges to better align visual and textual embeddings, mitigating the “soft positive sample” problem. We design an aggregated full-grained semantic representation learning network to resolve the “soft negative sample” issue arising from the lack of understanding of global information in the actual inference of traditional interactive embedding methods. This network incorporates an offline CLIP \cite{CLIP} model to generate global representations, enhancing global information comprehension and guiding the adaptive aggregation of local features, ultimately obtaining full-grained semantic embeddings. To fully explore high-order relationships between instances and enable the model to learn high-order shared knowledge, we model instance neighborhood relations by forming neighborhood subgraphs for each instance and its neighboring instances and introduce a graph neural network for modal interaction learning. Furthermore, inspired by MOCO \cite{MOCO}, we incorporate momentum encoders and a dynamic memory bank to expand the negative sample capacity, improving contrastive learning effects and feature discriminability.

By integrating these modules, our framework can thoroughly explore high-order associations and subtle differences among semantically similar instances at various stages, enhancing image-text retrieval performance while maintaining high efficiency.

Our main contributions can be summarized as follows:
\begin{itemize}
	\item
	To the best of our knowledge, we are the first to apply the clustering prototype contrastive learning method to the image-text matching task to address the “soft positive sample” problem. We perform cross-modal alignment in the clustering semantic space by capturing subtle semantic differences between samples through clustering mapping. This approach more effectively leverages the inherent semantic structure in the data.
	\item
	We propose a comprehensive framework for multi-granularity semantic representation learning. This framework integrates the advantages of coarse-grained and fine-grained retrieval methods, effectively capturing and utilizing multi-granularity semantic information.
	\item
	A method for instance neighborhood modeling is devised by us. It can learn richer shared knowledge by profoundly exploring high-order associations among instances, thereby obtaining superior overall embedding representations.
	\item
	Extensive experiments on several widely used public benchmarks for image-text matching (\textit{i.e}., Flickr30K, MSCOCO and ECCV Caption) demonstrate that our proposed framework achieves state-of-the-art performance.
\end{itemize}
\section{Related Work}\label{sec:RelatedWork}
\subsection{Image-Text Matching}
Image-text matching research falls into two main categories: global matching methods based on independent embeddings and local matching methods using interactive embeddings. Global matching methods represent images and text as holistic feature vectors in a shared semantic space. VSE++ \cite{VSE++} significantly improved visual semantic embeddings through hard negative sample mining, establishing a foundation for subsequent research.

\textbf{Global matching methods} have made significant progress in multiple aspects in recent years. In network architecture, DSRAN \cite{DSRAN} introduced a dual semantic relationship attention network for fragment-level and fragment-global semantic enhancement. VSRN++ \cite{VSRN++} proposed a visual semantic reasoning network to capture key objects and concepts. CORA \cite{CORA} utilized scene graphs to represent captions, employing graph attention networks for efficient matching. For feature aggregation, GPO \cite{GPO} introduced generalized pooling operators, adaptively learning optimal pooling functions for different modalities, achieving significant results in various
cross-modal tasks \cite{CORA,DCIN, USER, DADA, UAN}. ESA \cite{ESA} employed an external spatial attention aggregation module to enhance cross-modal representation. To improve fine-grained interaction, DIME \cite{DIME} proposed a dynamic multimodal interaction framework, while HREM \cite{HREM} introduced instance-level interaction for word-region correspondence learning. IMEB \cite{IMEB} incorporated memory networks to store interactive proxy features, balancing retrieval accuracy and efficiency.

\textbf{Local matching methods} focus on fine-grained cross-modal interactions and semantic alignment between image regions and text words, using these alignments to infer overall similarity. SCAN \cite{SCAN} introduced a stacked cross-attention model for fine-grained alignment, inspiring subsequent research. SGRAF \cite{SGRAF} employed multi-step reasoning through similarity graph reasoning and attention filtering modules to capture local-global alignment relationships. NAAF \cite{NAAF} adopted a dual-branch mechanism to measure both similarity and dissimilarity. CHAN \cite{CHAN} proposed a hard assignment encoding scheme to extract information-rich region-word pairs while eliminating redundant alignments. BOOM \cite{BOOM} introduced explicit bidirectional consistency constraints for more precise cross-modal semantic alignment. TVRN \cite{TVRN} enhances representation capability by integrating text and visual context information at the local level while leveraging the consistency of image-text features at the global level to guide semantic alignment.  Despite their excellent performance, these methods often face challenges in practical applications due to slow inference speeds.

The method proposed in this study belongs to the global matching category based on independent embeddings. However, it incorporates the cross-modal interaction advantages of local matching methods, which ensures accuracy while considering the high efficiency.
\subsection{Noise correspondence in VL datasets}
The current visual-language (VL) datasets \cite{COCO, f30k, LAION, CC3M, CC12M} commonly suffer from noisy correspondence issues, primarily including False positive samples and Soft positive samples. These problems become increasingly prominent as dataset scales expand.

\textbf{False positive samples} refer to noisy correspondences incorrectly labeled as positive samples, particularly prevalent in large-scale VL datasets \cite{LAION, CC3M,CC12M} constructed through web crawling. Currently, these are mainly filtered using CLIPScore \cite{Clipscore} or manual rules. Thao et al \cite{IMICAP}. address noise correction through Blip-2 \cite{blip2} caption synthesis, while Duan et al. \cite{pc2} propose the Pseudo-Classification based Pseudo-Captioning (PC\texorpdfstring{$^2$})) framework to achieve richer supervision for noisy correspondence pairs.

\textbf{Soft positive samples} (i.e., false negative samples) refer to positive relationships mislabeled as negative samples. Existing VL datasets \cite{COCO, f30k, LAION, CC3M, CC12M} only consider original image-caption pairs as positive samples, overlooking numerous potential positive correspondences. Chun et al. \cite{ECCV_Caption} constructed the ECCV Caption dataset through five state-of-the-art image-text matching models and human annotation. This dataset significantly expands the positive samples in MSCOCO, increasing the number of positive captions per image from 5 to 17.9 and positive images per caption from 1 to 8.5, providing a crucial benchmark for comprehensive retrieval evaluation. 
However, constructing the ECCV Caption requires substantial human and computational resources and is limited to covering only the MSCOCO test set. During the training phase, the issue of soft negative samples still confuses the model’s learning process. Li et al. \cite{LG} incorporated an external language pre-trained model during training to correct soft positive samples. FNE \cite{FNE} proposed a strategy to eliminate false negative samples by weighting the negative samples through Bayesian rules, reducing the negative impact of false negatives on model training. However, these methods either require additional computational resources or rely on dataset distribution. Unlike existing approaches, our proposed clustering prototype contrastive learning method dynamically learns unified prototype representations through online clustering, effectively leveraging the inherent semantic structure in the data without relying on annotations. This method effectively addresses the soft positive sample problem, eliminating the need for additional resources, and its effectiveness is validated through experiments on ECCV Caption dataset.
\subsection{Contrastive Learning}
\textbf{Self-Supervised Contrastive Learning} learns discriminative features by contrasting different data views without explicit labels, marking significant progress in visual representation learning. InstDisc \cite{InstDisc} pioneered instance discrimination as a pretext task, treating each instance as a distinct class. SimCLR \cite{SimCLR} enhanced this approach through data augmentation and demonstrated the benefits of larger batch sizes. To reduce computational overhead, MoCo \cite{MOCO} introduced momentum contrast with a dynamic dictionary. Building on these foundations, cluster-based methods emerged to leverage inherent semantic structural information in data: DeepCluster \cite{DeepCluster} utilized K-means clustering for pseudo-label generation, while SwAV \cite{SWAV} proposed online clustering with cross-view prediction.

\textbf{Image-Text Contrastive Learning} adapts multi-view contrastive principles \cite{CMC,InfoMin} to cross-modal scenarios. CLIP \cite{CLIP} achieved breakthrough performance through large-scale image-text pretraining. ALBEF \cite{ALBEF} incorporated momentum distillation for noise reduction, while SLIP \cite{SLIP} combined SimCLR and CLIP frameworks to leverage unimodal and cross-modal learning signals.
\section{The Proposed Method}\label{sec:The proposed method}
\begin{figure*}[t]
	\centering
	\includegraphics[width=\textwidth]{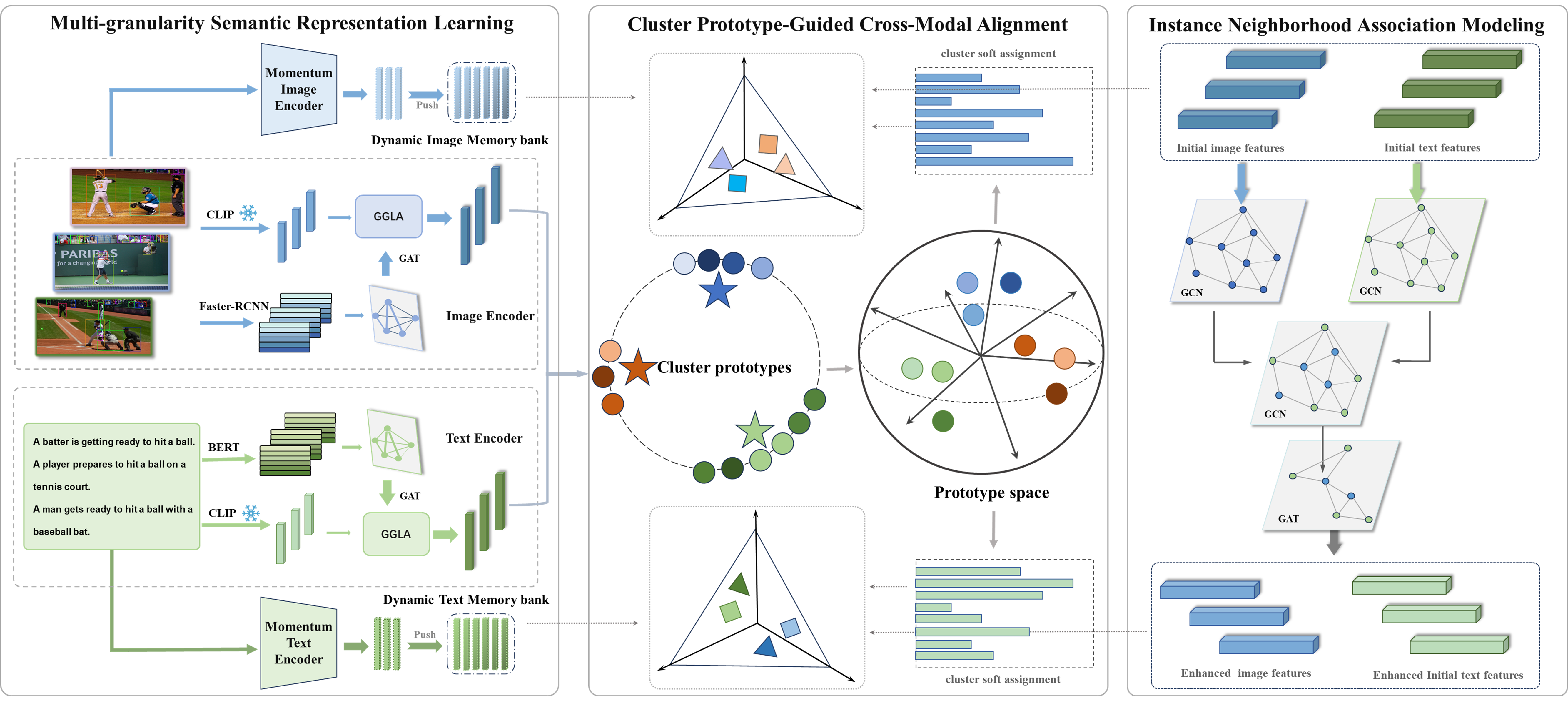}
	\caption{An overview of our AAHR framework, which comprises several key modules: \textbf{Multi-granularity Semantic Representation Learning} Network combining joint extraction of Global and Local features, Global semantics-Guided adaptive Local feature Aggregation(GGLA), and Full-Granularity feature gated Fusion; \textbf{Cluster Prototype-Guided Cross-Modal Alignment} module, which uses online clustering to serve as a bridge to align visual and textual embeddings; \textbf{Dynamic Memory Banks}, which enhance image-text contrastive learning by providing an expanded pool of negative samples; \textbf{Instance Neighborhood Semantic Association Modeling} module, which constructs correlation matrices and neighborhood subgraphs using global and local features, then applies graph neural networks for modal interaction learning, enabling deeper exploration of sample instance relationships.
	} \label{fig:Overview}
\end{figure*}
In this section, We will provide a detailed description of the proposed method, with its overall framework illustrated in Fig. \ref{fig:Overview}.
\subsection{Multi-granularity Semantic Representation Learning}
Existing image-text matching methods often have limitations: local feature-based approaches typically overlook aggregation of multi-level features, treating local features equally without considering global semantic guidance. Conversely, global feature-based methods often fail to perceive fine-grained local features. To address these issues, we propose a learning framework that integrates multi-granularity semantics.

This framework aggregates local features guided by global semantics, effectively mining and fusing multi-level semantic information from images and text. Through mutual feedback between global and local features, the global semantic representation gains support from fine-grained details, while local feature representation incorporates global semantic guidance. This approach enables effective semantic flow and complementarity across different levels.
\subsubsection{Joint Extraction of Global and Local Features}
As shown in Fig. \ref{fig:Overview}, for each input image $I$, we employ a bottom-up attention model \cite{BUTD} with a Faster R-CNN \cite{faster_r_cnn} network to extract the K=36 most salient region features. Through a fully connected linear projection layer, each region feature is mapped into a d-dimensional joint semantic space, resulting in image region representations $\hat{R}=\{\hat{\boldsymbol{r}}_1,\cdots,\hat{\boldsymbol{r}}_{n_r}\}\in\mathbb{R}^{n_r\times d}$, where $n_r$ denotes the number of region features.

For each input sentence $T$, a pre-trained BERT \cite{BERT} model extracts word-level feature representations. These are projected into the same d-dimensional joint semantic space using a fully connected linear projection layer, yielding $\hat{W}=\{\hat{\boldsymbol{w}}_1,\cdots,\hat{\boldsymbol{w}}_{n_t}\}\in\mathbb{R}^{n_t\times d}$, where $n_t$ represents the number of word-level features.

We enhance fine-grained semantic association features for both visual regions and textual words. For the visual modality, we construct a semantic relationship graph between visual areas and employ a graph attention network to learn contextual semantics. Graph nodes represent region features, with edges denoting semantic associations. Processing the original image region representations $\hat{R}=\{\hat{\boldsymbol{r}}_1,\cdots,\hat{\boldsymbol{r}}_{n_r}\}\in\mathbb{R}^{n_r\times d}$ through this network yields fine-grained semantic association-enhanced region features $R=\{{\boldsymbol{r}_1},\cdots,\boldsymbol{r}_{n_r}\}\in\mathbb{R}^{n_r\times d}$. Similarly, we derive enhanced word features $W=\{\boldsymbol{w}_1,\cdots,\boldsymbol{w}_{n_t}\}\in\mathbb{R}^{n_t\times d}$ for the textual modality.

To capture global semantic information, we utilize a pre-trained CLIP \cite{CLIP} model (ViT/B-32) without fine-tuning, considering computational resource limitations and to validate the effectiveness of our method. Through a fully connected linear projection layer, we extract global feature representations $\boldsymbol{v_g}\in\mathbb{R}^{d}$ and $\boldsymbol{t_g}\in\mathbb{R}^{d}$ for each image $I$ and sentence $T$, respectively.
\subsubsection{Global Semantics-Guided Adaptive Local Feature Aggregation}
Inspired by the encoding concept in image retrieval tasks \cite{Kernel_codebooks,soft_coding}, we propose an adaptive local feature pooling method guided by global semantics, which utilizes a soft assignment encoding framework to learn fine-grained feature pooling weights. For image encoding, we represent the salient regions of an image as $R=\{\boldsymbol{r}_1,\cdots,\boldsymbol{r}_{n_r}\}\in\mathbb{R}^{n_r\times d}$, and treat these regions as the image's codebook. Each salient region feature vector $\boldsymbol{r}_i$ ($i \in [1, n_r]$) is considered a codeword.

Using the CLIP-extracted image global feature vector $\boldsymbol{v_g} \in \mathbb{R}^{d}$ as the query, we calculate the soft assignment encoding coefficient vector $\boldsymbol{a^v} \in \mathbb{R}^{n_r}$ concerning codebook $R$.
\begin{equation}
	\boldsymbol{a_j^v}=\frac{\boldsymbol{v_g}^\top{\boldsymbol{r}_j}}{\|\boldsymbol{v_g}\|\cdot\|{\boldsymbol{r}_j}\|},
\end{equation}
where $\boldsymbol{a_j^v}$ represents the membership of $\boldsymbol{v_g}$ (query) on the $j$-th salient region feature $\boldsymbol{r}_{j}$.

To capture diverse patterns, we apply multiple linear transformations to the query vector and codebook, obtaining a query set $Q^v=\{{\boldsymbol{q}_1^v,\cdots,\boldsymbol{q}_m^v}\}$ and a codebook set $C^v=\{{C_1^v,\cdots,C_m^v}\}$, where $m$ is the number of transformations. Then, for the $k$-th transformation, we calculate its soft assignment encoding coefficient vector $\boldsymbol{a}_k^v \in \mathbb{R}^{n_r}$ as follows:
\begin{equation}
	\boldsymbol{a}_{k,j}^v=\frac{{\boldsymbol{q}_k^v}^{\top}\boldsymbol{c}_{k,j}^v}{\|\boldsymbol{q}_k^v\|\cdot\|\boldsymbol{c}_{k,j}^v\|},
\end{equation}
where $\boldsymbol{c}_{k,j}^v$ is the $j$-th codeword in the codebook $C_k^v$.

We then perform mean pooling on all soft assignment encoding coefficient vectors to obtain the final vector $\boldsymbol{a}^V \in \mathbb{R}^{n_r}$:
\begin{equation}
	\boldsymbol{a}^V=\frac{1}{m}\sum_{k=1}^m\boldsymbol{a}_{k}^v,
\end{equation}

To convert these coefficient vectors into fine-grained feature pooling weights, we apply softmax to $\boldsymbol{a}^V$:
\begin{equation}
	{w}^v=softmax(\boldsymbol{a}^V),
\end{equation}
where ${w}^v\in \mathbb{R}^{n_r}$ represents the learned fine-grained feature pooling weights for the image. Similarly, we can obtain the pooling weights ${w}^t\in \mathbb{R}^{n_t}$ for the text. We then perform weighted aggregation of the fine-grained features:
\begin{equation}
	\begin{aligned}
		\boldsymbol{v_f} = \sum_{i=1}^{n_r} w^v_i \boldsymbol{r}_i,\\
		\boldsymbol{t_f} = \sum_{j=1}^{n_t} w^t_j \boldsymbol{w}_j,
	\end{aligned}
\end{equation}
where $\boldsymbol{v_f} \in \mathbb{R}^d$ and $\boldsymbol{t_f} \in \mathbb{R}^d$ are the fine-grained aggregated feature representations for the image and text, respectively.
\subsubsection{Full-Granularity Feature Gated Fusion}
To obtain full-granularity aggregated features, we fuse the fine-grained aggregated features $\boldsymbol{v_f}$ and $\boldsymbol{t_f}$ with the global feature representations $\boldsymbol{v_g}$ and $\boldsymbol{t_g}$. Instead of manually adjusted scaling hyperparameters, we employ a gated fusion mechanism that allows the model to learn weighting coefficients automatically through trainable weights.
For the image full-granularity aggregated feature $\boldsymbol{v}$, we use:
\begin{equation}
	\begin{gathered}\boldsymbol{v} ={g}\cdot\boldsymbol{v_f}+(1-g) \cdot \boldsymbol{v_g},  \\g =\operatorname{sigmoid}(\mathbf{W}_v[\boldsymbol{v}_f,\boldsymbol{v}_g]+b_v),
	\end{gathered}
\end{equation}
where $g$ is a gating value used to balance the importance of $\boldsymbol{v_f}$  adaptively and $\boldsymbol{v}_g$, and $[\cdot ,\cdot]$ represents the concatenation operation. Similarly, for the text full-granularity aggregated feature $\boldsymbol{t}$:
\begin{equation}
	\begin{gathered}\boldsymbol{t} ={h}\cdot\boldsymbol{t_f}+(1-h) \cdot \boldsymbol{t_g},  \\h =\operatorname{sigmoid}(\mathbf{W}_t[\boldsymbol{t}_f,\boldsymbol{t}_g]+b_t).\end{gathered}
\end{equation}
This approach enables adaptive fusion of fine-grained and global features, resulting in full-granularity aggregated feature representations for both image and text modalities.
\begin{figure*}[t]
	\centering
	\includegraphics[width=\textwidth]{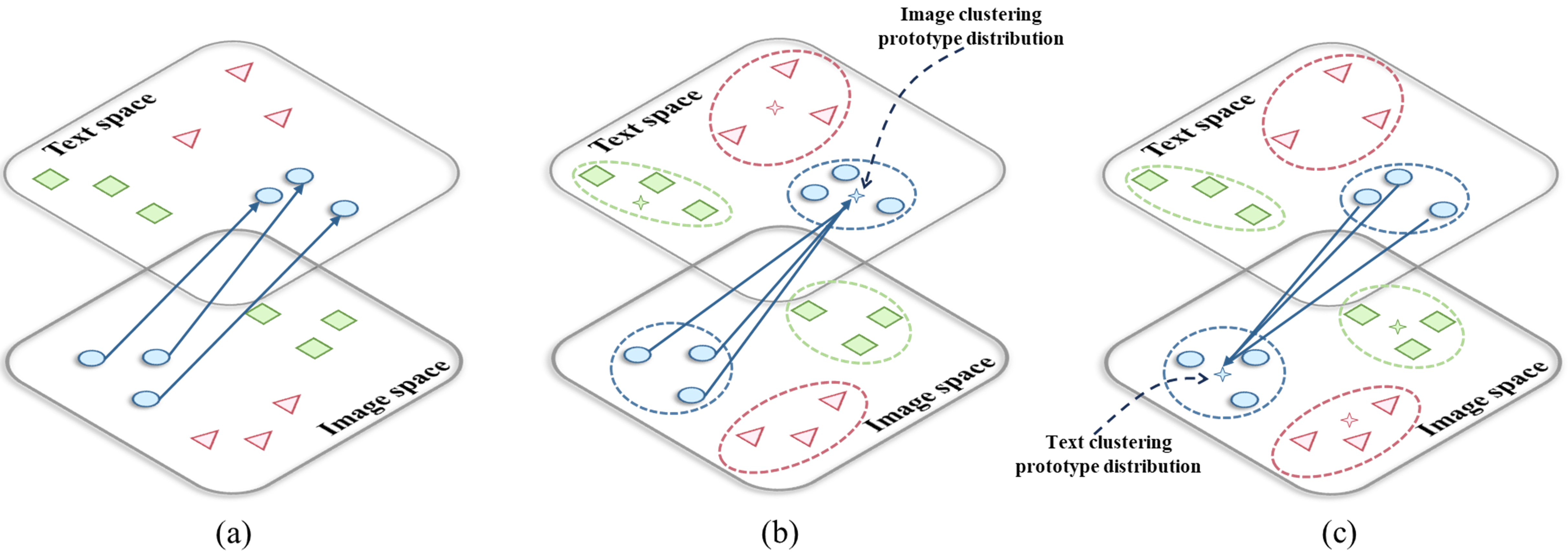}
	\caption{Illustration of cluster Prototype-Guided cross-modal Alignment (PGA): (a) Instance-based image-text contrastive learning aims to draw positive pairs together while separating negative pairs. (b \& c) Cluster Prototype-Guided Cross-modal Alignment achieves precise alignment in the prototype space by computing bidirectional cross-modal prototype distribution matrices. This approach particularly benefits “soft positive samples” that instance discrimination struggles to distinguish. Leveraging similar prototype distributions enhances image-text alignment for these cases, thereby improving semantic structural consistency.
	} \label{fig:PGA}
\end{figure*}
\subsection{Cluster Prototype-Guided Cross-Modal Alignment}
Previous image-text matching tasks have predominantly employed self-supervised contrastive learning frameworks \cite{GPO,HREM,CHAN,CLIP,ALIGN,FLIP}, utilizing objective functions such as InfoNCE loss \cite{InfoNCE} and Hard Negative Triplet loss \cite{VSE++}. However, these methods may inadvertently treat semantically similar sample pairs as negative examples during random negative sampling, potentially causing the model to learn incorrect associations. This issue is particularly pronounced for Hard Negative Triplet loss, where selected hard negative examples may be semantically close to positive samples, leading to learning confusion.

As illustrated in Fig. \ref{fig:motivation}(b), semantically similar pairs can be classified as positive and negative examples, resulting in embedding vectors that are difficult to distinguish. If mistakenly viewed as negative samples, these pairs may interfere with the model's learning process.

Traditional image-text contrastive learning methods typically employ instance discrimination as an auxiliary pretext task, treating the image and text of each instance as an independent category, as shown in Fig. \ref{fig:PGA}(a). However, this approach often includes numerous “soft positive samples” in the constructed negative sample set, potentially leading to model confusion.

To address this issue, we innovatively designed a cluster prototype-guided cross-modal alignment module, as depicted in Fig. \ref{fig:PGA}(b) and Fig. \ref{fig:PGA}(c). This approach first captures subtle semantic differences between samples through cluster mapping, followed by cross-modal alignment in the clustered semantic space. This method more effectively leverages the inherent semantic structural information within the data.

Specifically, we first define a set of trainable prototype vectors $\boldsymbol{P}=\{\boldsymbol{p}_1, \boldsymbol{p}_2, \cdots, \boldsymbol{p}_k\}$ in the joint image-text representation space, where $k$ is the number of prototypes. These vectors represent low-dimensional projections of the dataset, storing frequent, general, and transferable features. For image embeddings $\boldsymbol{V}=\{\boldsymbol{v}_1,\cdots,\boldsymbol{v}_m\}$ and text embeddings $\boldsymbol{T}=\{\boldsymbol{t}_1,\cdots,\boldsymbol{t}_m\}$, we project them into a unified prototype space to obtain similarity scores:
\begin{equation}
	\begin{gathered}
		\boldsymbol{u}^{v}=\operatorname{softmax}(\boldsymbol{V}^\top \boldsymbol{P})\\\boldsymbol{u}^{t}=\operatorname{softmax}(\boldsymbol{T}^\top \boldsymbol{P})\\
	\end{gathered},
\end{equation}
where $\boldsymbol{u}^{v}, \boldsymbol{u}^t\in \mathbb{R}^{m \times k}$, with each row representing the similarity scores between a sample and $k$ prototypes. We then use the Sinkhorn algorithm \cite{Sinkhorn} to obtain soft assignment matrices $\boldsymbol{D}^v, \boldsymbol{D}^t\in\mathbb{R}^{m \times k}$ for images and texts from the similarity scores, ensuring that each sample is softly assigned to multiple prototypes, which can better capture subtle semantic
differences between data than hard assignments.

Then, we calculate the Prototype-Guided cross-modal Alignment (PGA) loss from both image and text perspectives:

From the image perspective:
\begin{equation}
	\label{PGA_img}
	\mathcal{L}_{img}^{PGA}=-\frac1m\sum_{i=1}^m\sum_{j=1}^k\boldsymbol{D}^t_{ij}\log\frac{\exp(\boldsymbol{u}^v_{ij}/\tau)}{\sum_{l=1}^k\exp(\boldsymbol{u}^v_{il}/\tau)},
\end{equation}

From the text perspective:
\begin{equation}
	\mathcal{L}_{txt}^{PGA}=-\frac1m\sum_{i=1}^m\sum_{j=1}^k\boldsymbol{D}^v_{ij}\log\frac{\exp(\boldsymbol{u}^t_{ij}/\tau)}{\sum_{l=1}^k\exp(\boldsymbol{u}^t_{il}/\tau)}
	\label{PGA_txt},
\end{equation}
where $m$ is the batch size, $k$ is the number of prototypes, $\tau$ is the temperature coefficient, $\boldsymbol{D}^t_{ij}$ and $\boldsymbol{D}^v_{ij}$ represent the assignment probabilities, and $\boldsymbol{u}^v_{ij}$ and $\boldsymbol{u}^t_{ij}$ denote the similarity scores. 
The final PGA loss is:
\begin{equation}
	\mathcal{L}^{PGA}=\mathcal{L}^{PGA}_{img}+\mathcal{L}^{PGA}_{txt}.
\end{equation}

This prototype-based contrastive learning approach better utilizes the inherent semantic structure of the data, achieving finer-grained alignment in the prototype space. It effectively mitigates the “soft positive sample” problem by not directly relying on image-text annotation information. When the soft assignment probability of a text (or image) to a specific prototype is high, the similarity score between the corresponding image (or text) and that prototype is encouraged to increase, and vice versa. This mechanism captures semantic relationships between cross-modal data, thereby enhancing the cross-modal alignment capability of the model.
\subsection{Dynamic Memory Bank-Assisted Image-Text Contrastive Learning}
Existing image-text matching methods often limit training to samples within each mini-batch, restricting the scale of negative samples and constraining the model's representational capacity. Inspired by MoCo \cite{MOCO}, we introduce dynamic memory banks for both image and text modalities to store a large number of negative samples. This approach decouples memory bank capacity from mini-batch size, allowing us to scale negative samples beyond traditional mini-batch training constraints, thus enhancing the image-text contrastive learning process and overall model performance.

Specifically, let $\theta^v$ and $\theta^t$ denote the parameters of the visual encoder $f^v$ and the text encoder $f^t$, respectively. We introduce momentum encoders $f^v_m$ (with parameters $\theta^v_m$) and $f^t_m$ (with parameters $\theta^t_m$) for the visual and text modalities. Their parameters are updated based on momentum:
\begin{equation}
	\begin{aligned}
		{\theta}^{v}_m &= \tilde m\cdot{\theta}^v_m+(1-\tilde m)\cdot\theta^v,\\
		{\theta}^{t}_m &= \tilde m\cdot{\theta}^t_m+(1-\tilde m)\cdot\theta^t,
	\end{aligned}
\end{equation}
where $\tilde m$ represents the momentum coefficient. To maintain consistency of representations in the dynamic memory bank and ensure smooth feature updates, $\tilde m$ is typically set to a relatively high value. In this paper, we set $\tilde m = 0.999$.

We establish two dynamic memory banks $B_v=\{\boldsymbol{z^v_1},\boldsymbol{z^v_2},\cdots,\boldsymbol{z^v_{N_{v}}}\}$ and $B_t=\{\boldsymbol{z^t_1},\boldsymbol{z^t_2},\cdots,\boldsymbol{z^t_{N_t}}\}$ to store features extracted by the visual momentum encoder $\boldsymbol{z^v_j}, j\in[1,N_v]$ and the text momentum encoder $\boldsymbol{z^t_j}, j\in[1,N_t]$, respectively. $N_v$ and $N_t$ denote the sizes of the visual and text dynamic memory banks. During each training iteration, the current batch data is processed by momentum encoders, and the resulting representations are added to the memory banks while removing the oldest batch.

To fully utilize the large number of negative samples we have mined, we adopt the InfoNCE loss \cite{InfoNCE} as our objective function. The momentum contrastive loss (MCL) from image to text is defined as:
\begin{equation}
	\mathcal{L}^{MCL}_{img}=-\sum_{i=1}^{m}\log\frac{\exp \left(\text {sim}(\boldsymbol{v}_i ^{\top},\boldsymbol{z}^t_i)/\tau\right)}{\sum_{j=1}^{N_t}\exp \left(\text {sim}(\boldsymbol{v}_i ^{\top},\boldsymbol{z}^t_j)/\tau\right)},
\end{equation}
where $\boldsymbol{v}_i$ is the feature vector of the current batch image, and $\boldsymbol{z}^t_i$ is the momentum feature vector of its paired text. The similarity function $\text{sim}(\cdot,\cdot)$ uses cosine similarity.

Similarly, the MCL from text to image is:
\begin{equation}
	\mathcal{L}^{MCL}_{txt}=-\sum_{i=1}^{m}\log\frac{\exp \left(\text {sim}(\boldsymbol{t}_i ^{\top},\boldsymbol{z}^v_i)/\tau\right)}{\sum_{j=1}^{N_v}\exp \left(\text {sim}(\boldsymbol{t}_i ^{\top},\boldsymbol{z}^v_j)/\tau\right)},
\end{equation}
where $\boldsymbol{t}_i$ is the feature vector of the current batch text, and $\boldsymbol{z}^v_i$ is the momentum feature vector of its paired image.

The final momentum contrastive loss is defined as:
\begin{equation}
	\mathcal{L}^{MCL}=\mathcal{L}^{MCL}_{img} + \mathcal{L}^{MCL}_{txt}.
\end{equation}
\subsection{Instance Neighborhood Semantic Association Modeling}
Samples within the same batch often share rich semantic information. To effectively harness this information and obtain discriminative visual semantic embedding, we integrate shared semantic knowledge into cross-modal feature interactions. We construct a multimodal instance neighborhood semantic association graph and apply graph neural networks for intra-modal and inter-modal semantic interactions.
\subsubsection{Multimodal Instance Neighborhood Semantic Association Graph}
We construct intra-modal and inter-modal instance neighborhood semantic association graphs separately. For the intra-modal association graph, we use the Euclidean distance of full-granularity aggregated visual(textual) embeddings to quantify the semantic similarity between instances:
\begin{equation}
	\boldsymbol S^{I\to I}_{ij}=e^{-\frac{\|\boldsymbol{v}_i-\boldsymbol{v}_j\|_2^2}{\varepsilon }}, \boldsymbol  S^{T\to T}_{ij}=e^{-\frac{\|\boldsymbol{t}_i-\boldsymbol{t}_j\|_2^2}{\varepsilon }}.
\end{equation}
Here, $\boldsymbol{v}_i, \boldsymbol{v}_j \in \mathbb{R}^d$ denote the full-granularity aggregated feature vectors for the  $i$-th and $j$-th images, and $\boldsymbol{t}_i, \boldsymbol{t}_j \in \mathbb{R}^d$ represent those for the corresponding texts, $\|\cdot\|_2^2$ represents the $\ell_2$ norm, and $\varepsilon$ is a temperature parameter.

Due to the semantic gap between image and text, we adopt fine-grained matching of local features to measure cross-modal instance neighborhood semantic associations. Specifically, for an image-text pair and their region-word similarity matrix, we find the text word (or visual region) with the highest matching score for each region (or each word) and average these maximum matching scores:
\begin{equation}
	\begin{split}\boldsymbol S_{ij}^{I\to T} &=\frac{1}{n_r}\sum_{\tau=1}^{n_r}\max_{\eta\in[1,n_t]}\left(\boldsymbol{r}_{i,\tau}^\top\boldsymbol{w}_{j,\eta }\right),\\\boldsymbol S_{ij}^{T\to I} &=\frac{1}{n_t}\sum_{\tau=1}^{n_t}\max_{\eta\in[1,n_r]}\left(\boldsymbol{w}_{i,\tau}^{\top}\boldsymbol{r}_{j,\eta}\right),\end{split}
\end{equation}
where $n_r$ and $n_t$ represent the number of image regions and text words respectively, $\boldsymbol{r}_{i,\tau}, \boldsymbol{w}_{j,\eta} \in \mathbb{R}^d$ denote the $\tau$-th region feature of the $i$-th image and the $\eta$-th word feature of the $j$-th text, respectively.

Finally, the resulting multimodal instance neighborhood semantic association graph can be represented as:
\begin{equation}
	\boldsymbol{S}=\begin{bmatrix}\boldsymbol S^{I\to I}&\boldsymbol S^{I\to T}\\\boldsymbol S^{T\to I}&\boldsymbol S^{T\to T}\end{bmatrix}\in\mathbb{R}^{2m\times2m},
\end{equation}
where $m$ is the batch size.
\subsubsection{Instance Neighborhood Semantic Interaction}
After constructing the multimodal instance neighborhood semantic association graph, we apply a multimodal graph neural network on this graph to perform instance neighborhood semantic interactions, capturing higher-order semantic relationships between instances.

Specifically, we introduce Graph Convolutional Networks (GCN) to achieve intra-modal neighborhood semantic interactions in both image and text modalities. For the image modality, the $l$-th layer of GCN can be represented as:
\begin{equation}
	\boldsymbol{H}_I^{(l)}=\sigma\left(\tilde{\boldsymbol{D}_I}^{-\frac{1}{2}}\boldsymbol S^{I\to I}\tilde{\boldsymbol{D}_I}^{-\frac{1}{2}}\boldsymbol{H}_I^{(l-1)}\boldsymbol{W}_I^{(l)}\right),
\end{equation}
where $\boldsymbol{H}_I^{(l)} \in \mathbb{R}^{m\times d}$ denotes the image node feature matrix at the $l$-th layer, $S^{I\to I} \in \mathbb{R}^{m \times m}$ represents the image modality instance neighborhood semantic association graph, $\boldsymbol{W}_I^{(l)} \in \mathbb{R}^{d\times d}$ is the learnable weight matrix for the $l$-th layer, $\sigma(\cdot)$ is a non-linear activation function (we use LeakyReLU), and $\tilde{\boldsymbol{D}_I} \in \mathbb{R}^{m \times m}$ is the diagonal element of the degree matrix.

Similarly, for the text modality:
\begin{equation}
	\boldsymbol{H}_T^{(l)}=\sigma\left(\tilde{\boldsymbol{D}_T}^{-\frac{1}{2}}\boldsymbol S^{T\to T}\tilde{\boldsymbol{D}_T}^{-\frac{1}{2}}\boldsymbol{H}_T^{(l-1)}\boldsymbol{W}_T^{(l)}\right).
\end{equation}
To further capture higher-order inter-modal neighborhood semantic associations, we employ a joint-modal GCN to aggregate neighborhood structural knowledge. Specifically, we concatenate the outputs of the image and text modality GCNs along the batch dimension and then input the concatenated features into the joint-modal GCN for cross-modal interaction:
\begin{equation}
	\begin{aligned}
		\boldsymbol{H}^{(l)}&=\sigma\left(\tilde{\boldsymbol{D}}^{-\frac{1}{2}}\boldsymbol S \tilde{\boldsymbol{D}}^{-\frac{1}{2}}\boldsymbol{H}^{(l-1)}\boldsymbol{W}^{(l)}\right)\\
		&=\sigma\left(\tilde{\boldsymbol{D}}^{-\frac{1}{2}}\begin{bmatrix}\boldsymbol{S}^{I\to I} &\boldsymbol{S}^{I\to T}\\\boldsymbol{S}^{T\to I}&\boldsymbol{S}^{T\to T}\end{bmatrix}\tilde{\boldsymbol{D}}^{-\frac{1}{2}} \begin{bmatrix} \boldsymbol{H}_I^{(l-1)} \\ \boldsymbol{H}_T^{(l-1)} \end{bmatrix}\boldsymbol{W}^{(l)}\right),
	\end{aligned}
\end{equation}
where $\boldsymbol{H}^{(l)} \in \mathbb{R}^{2m \times d}$ is the joint-modal node feature matrix at the $l$-th layer, $\boldsymbol{S} \in \mathbb{R}^{2m \times 2m}$ is the multimodal instance neighborhood semantic association graph.

We apply a joint-modal Graph Attention Network (GAT) to adaptively and finely capture neighborhood semantic associations to avoid over-smoothing phenomena and noisy semantic associations. The joint-modal GCN output serves as input to the GAT, which assigns different attention weights to neighbors, mining semantic associations more precisely.

To improve the quality of instance neighborhood semantic associations, we optimize the original multimodal instance neighborhood association graph $\boldsymbol{S}$ to obtain an optimized adjacency matrix $\tilde{\boldsymbol{S}}$. Specifically, we sort the element values in $\boldsymbol{S}$. For each submatrix $S^{I\to I}$, $S^{I\to T}$, $S^{T\to I}$, and $S^{T\to T}$, we set the elements ranked in the bottom 50\% to $-\infty$, while multiplying the top 50\% by an amplification coefficient $\alpha$:
\begin{equation}
	\label{at}
	\tilde{\boldsymbol{S}}_{ij}^{X\to Y} =
	\begin{cases}
		\alpha  \boldsymbol{S}_{ij}^{X\to Y}, & \mathrm{if~}\boldsymbol{S}_{ij}^{X\to Y}\geq\Lambda(\boldsymbol{S}^{X\to Y}) \\
		-\infty, & \mathrm{otherwise}
	\end{cases} \\,
\end{equation}
where $X, Y \in \{I, T\}$, and ${\Lambda}(\cdot)$ represents the median operation. This preserves strong semantic associations while filtering out noisy ones.

After obtaining the optimized adjacency matrix $\tilde{\boldsymbol{S}}$, we incorporate it into the self-attention score calculation of GAT:
\begin{equation}
	\operatorname{Att}(\boldsymbol{Q},\boldsymbol{K},\boldsymbol{V};\tilde{\boldsymbol{S}})=\operatorname{softmax}\left(\frac{\boldsymbol{Q}\boldsymbol{K}^T}{\sqrt{d_k}}+\tilde{\boldsymbol{S}}\right)\boldsymbol{V},
\end{equation}
where $\boldsymbol{Q}, \boldsymbol{K}, \boldsymbol{V} \in \mathbb{R}^{2m \times d}$ are query, key, and value matrices obtained by linear transformation of the joint-modal GCN output, and $d_k$ is the dimension of the key vector.

This method effectively incorporates instance neighborhood semantic association information into the feature learning process, enhancing the model's ability to capture complex semantic relationships. After instance neighborhood semantic interaction, we obtain the final enhanced visual semantic embeddings, denoted as $\{\boldsymbol{\hat v_i}\}_{i=1}^m$ and $\{\boldsymbol{\hat t_i}\}_{i=1}^m$, where $m$ is the batch size.

To accelerate convergence and improve embedding quality, we adopt the hard negative triplet loss \cite{VSE++} $\mathcal{L}^{tri}$:
\begin{equation}
	\mathcal{L}_\text{tri}(\boldsymbol v,\boldsymbol t)=[\gamma-s(\boldsymbol v,\boldsymbol t)+s(\boldsymbol v,\boldsymbol t^-)]_++[\gamma-s(\boldsymbol v,\boldsymbol t)+s(\boldsymbol v^-,\boldsymbol t)]_+,
\end{equation}
where $\gamma$ represents the margin hyperparameter,  $(\boldsymbol v,\boldsymbol t)$ is a positive image-text pair, $\boldsymbol{v}^-=\operatorname{argmax}_{x\neq\boldsymbol{v}}s(x,\boldsymbol{t})$ and $\boldsymbol{t}^-=\operatorname{argmax}_{y\neq\boldsymbol{t}}s(\boldsymbol{v},y)$ are the most difficult negative image and text in the batch, respectively.

To maintain the speed advantage of the dual-stream encoder during inference, we compute the triplet loss for both the original visual semantic embeddings and the enhanced embeddings, then perform cross-alignment. Therefore, the total instance neighborhood semantic interaction loss is defined as:
\begin{equation}
	\mathcal{L}^{NSI} = \mathcal{L}_{tri}(\boldsymbol{v},\boldsymbol{t}) + \mathcal{L}_{tri}(\boldsymbol{\hat{v}},\boldsymbol{\hat{t}}) + \mathcal{L}_{tri}(\boldsymbol{v},\boldsymbol{\hat{t}}) + \mathcal{L}_{tri}(\boldsymbol{\hat{v}},\boldsymbol{t}) + \mathcal{L}^{PGA}(\boldsymbol{\hat{v}},\boldsymbol{\hat{t}}),
\end{equation}
where $\mathcal{L}^{PGA}(\boldsymbol{\hat v},\boldsymbol{\hat{t}})$ represents the prototype-guided cross-modal alignment loss computed on the enhanced embeddings, which helps mitigate potential issues caused by “soft positive samples” and further improves the model's performance.
\subsection{Objective Function}
The final objective function is:
\begin{equation}
	\mathcal{L}=\mathcal{L}^{PGA}+\mathcal{L}^{MCL}+\mathcal{L}^{NSI}.
\end{equation}
\subsection{Discussions}
In this section, we elaborate on our research motivation and technical contributions.

\textbf{Motivation. }The challenge of similar instances in image-text matching tasks has been a fundamental yet long-overlooked issue. These semantically similar instances manifest in different forms throughout the model's training and inference lifecycle. Our analysis reveals three dimensions of this problem and their intrinsic connections:
\begin{enumerate}
\item 
The soft positive sample issue exists at the dataset construction. Current vision-language datasets \cite{COCO, f30k, LAION, CC3M, CC12M} employ strict image-text pair annotations, labeling semantically similar but unpaired samples as negative instances. The limitations of annotation result in numerous potential positive samples being mislabeled, affecting both model training effectiveness and evaluation accuracy.
\item 
The soft negative sample issue emerges during model inference. Traditional local matching methods overly emphasize detail alignment while neglecting global semantics, leading to erroneous matches that are "locally similar but semantically contradictory." This highlights the importance of balancing local details and global semantics in cross-modal matching. 
\item 
The challenge of effectively utilizing shared knowledge between similar instances. Observations reveal that semantically similar instances possess higher-order correlations. For example, as shown in Fig. \ref{fig:motivation}(a), "preparation for hitting" poses in different sports scenarios, despite varying in specific details, essentially convey similar semantic concepts. Mining this shared knowledge is crucial for enhancing the model's generalization capabilities.
\end{enumerate}

While these issues manifest differently, they fundamentally stem from the challenge of properly handling and utilizing similar instances in the data. The mislabeling of soft positive samples originates from the static perspective of dataset construction, soft negative samples expose limitations in model alignment mechanisms, and the lack of neighborhood relationship modeling hinders effective semantic knowledge transfer. Existing methods fail to exploit the inherent semantic topological structure within the data fully, lacking a unified framework to address these challenges. This prompts us to reconsider: How can we design a unified framework that accurately identifies potential positive relationships, effectively distinguishes true negative samples, while fully utilizing semantic connections between instances?

\textbf{Technical Contribution. }
To systematically address these three challenges, we propose the AAHR framework, which forms a unified framework through three modules:
\begin{enumerate}
	\item 
	To address the soft positive sample issue, we innovatively introduce prototype-based contrastive learning into image-text matching tasks. Traditional instance-discriminative contrastive learning treats each sample as an independent class, which proves too rigid when dealing with similar samples. By dynamically learning unified prototypes in the joint representation space, we establish a flexible semantic alignment mechanism. The soft assignment strategy in the prototype space enables the model to capture subtle semantic differences between samples, providing an elegant solution for handling soft positive samples. This approach no longer solely relies on original image-text annotations but fully utilizes the inherent semantic structure of the data.
	\item 
	To tackle the soft negative sample issue, we design a multi-granularity semantic representation learning framework. By incorporating CLIP \cite{CLIP} to generate global semantic representations, it provides effective guidance for local feature aggregation. This design preserves the discriminative power of fine-grained features while ensuring global semantic consistency.
	\item 
	For instance neighborhood semantic modeling, we propose an solution based on graph neural networks. The technical contributions of this module are twofold: 1) constructing a multimodal instance neighborhood semantic association graph provides structured support for mining higher-order shared knowledge; 2) combining a dynamic memory bank expands negative sample capacity, enhancing feature discriminability. This design effectively captures semantic associations between instances and promotes knowledge transfer and sharing through the message-passing mechanism of the graph structure.
\end{enumerate}

The organic integration of these technical modules forms a comprehensive solution that not only addresses the challenges of soft positive samples, soft negative samples, and neighborhood modeling individually, but they work collaboratively and complement each other: prototype-based learning provides more stable semantic references for multi-granularity features, while graph-structured neighborhood modeling further enhances the expressive power of the prototype space. Extensive experiments in the following section \ref{sec:Experiments} validate the superior performance of this framework across multiple benchmarks.

\section{Experiments}\label{sec:Experiments}
\subsection{Datasets and Evaluation Metrics}
\subsubsection{Datasets}
This study employs several widely used benchmark datasets for cross-modal image-text matching tasks: Flickr30K \cite{f30k}, MSCOCO \cite{COCO}, and ECCV Caption \cite{ECCV_Caption}.

Flickr30K dataset comprises 31,783 images, each with five human-annotated descriptive sentences. Following the standard splitting protocol \cite{splittingprotocol}, we use 29,000 images for training, 1,000 for validation, and 1,000 for testing.

MSCOCO contains 123,287 images, each paired with five descriptive texts. In line with previous studies \cite{GPO, CHAN}, we divide the dataset as follows: 113,287 images for training, 5,000 for validation, and 5,000 for testing. We report results for two test settings: (1) 5K test set: direct evaluation on all 5,000 test images; (2) 1K test set: testing on five subsets of 1,000 images each and averaging the results.

ECCV Caption dataset is a extension to the MSCOCO test set, addressing the soft-positive sample issue prevalent in existing image-text datasets. The original MSCOCO test set contains 5,000 images, each accompanied by 5 captions, totaling 25,000 captions, with each caption matched to only one image. This matching scheme overlooks many potential semantic associations, causing many semantically relevant image-text pairs to be incorrectly classified as negative samples (i.e., soft positives). To address this issue, ECCV Caption re-annotated a subsampled dataset of 1,261 images and 1,332 captions, increasing the number of correct captions per image to an average of 17.9 (compared to 5 per image in the original MSCOCO) and the number of relevant images per caption to an average of 8.5 (compared to 1 per caption in the original). With this dense semantic annotation, the coverage of the dataset has been significantly improved, providing a more comprehensive and reliable benchmark for evaluating model performance, particularly in soft positive sample recall.
\subsubsection{Evaluation Metrics}
We employ multiple complementary metrics to assess model performance comprehensively. For the Flickr30K and MSCOCO datasets, we utilize the standard Recall@K (R@K) as the basic metric to measure the proportion of correct matches found within the top-K retrieval results. The R@K metric is formally defined as:
\begin{equation}
	R@K = \frac{|G \cap S_K|}{|G|},
\end{equation}
where G denotes the set of ground truth matches and $S_K$ represents the set of top-K retrieved results.
Additionally, we incorporate rSum to reflect the overall performance in bi-directional image-text retrieval tasks:
\begin{equation}
	rSum=\underbrace{R@1+R@5+R@10}_{image-to-text}+\underbrace{R@1+R@5+R@10}_{text-to-image}.
\end{equation}

To more accurately evaluate the model's capability in handling complex matching relationships, we adopt the mAP@R and R-P metrics introduced by Chun et al. \cite{ECCV_Caption} for the ECCV Caption dataset. The R-P metric evaluates precision at rank position R:
\begin{equation}
R\text{-}P = \frac{|G \cap S_R|}{R},
\end{equation}
where R = |G| is the number of ground truth matches.
The mAP@R metric calculates the mean Average Precision across all queries until all ground truth matches are found:
\begin{equation}
	mAP@R=\frac{1}{|Q|}\sum_{q\in Q}\frac{1}{|G_q|}\sum_{k=1}^{n_q}P(k)\times rel(k),
\end{equation}
where Q is the set of all queries, $G_q$ is the set of ground truth matches for query q, $n_q$ is the position of the last ground truth match, P(k) represents the precision at cutoff k, and rel(k) is an indicator function equaling one if the item at rank k is relevant and zero otherwise. Compared to R@K, which only focuses on top-K results, mAP@R provides a more comprehensive evaluation by considering the ranking quality until all relevant items are retrieved. Human evaluation studies \cite{ECCV_Caption} have demonstrated that these new metrics align better with real-world application scenarios and better reflect the model's ability to identify and recall soft positive samples.
\subsection{Implementation Details}
We employ a Faster R-CNN \cite{faster_r_cnn} model pre-trained on the Visual Genome dataset \cite{Visualgenome} (based on a ResNet-101 \cite{resnet} backbone) to extract image region features. Specifically, we select the K = 36 region proposals with the highest confidence scores, each represented by a 2,048-dimensional feature vector. Concurrently, for text, we employ a pre-trained BERT-base model \cite{BERT} to obtain 768-dimensional word-level features. Additionally, we use CLIP (ViT/B-32) \cite{CLIP} to extract 512-dimensional global representations for both images and text, enhancing global semantic understanding. The CLIP model is not fine-tuned in our experiments.

We utilize the AdamW optimizer with an initial learning rate of 5e-4. The batch size is 128 for Flickr30K and 256 for MSCOCO. The joint embedding space dimension is set to 1,024, the momentum coefficient $\tilde m$ to 0.999, the amplification factor $\alpha$ to 1.5, and the margin hyperparameter $\gamma$ to 0.2. The temperature hyperparameter $\tau$ is set to 0.1, and the balancing factor $\varepsilon$ is set to 1. For the graph neural networks, we use single-layer architectures with a dropout rate of 0.6 for GCN and 0.1 for GAT. The dynamic memory bank sizes are 2,048 for Flickr30K and 4,096 for MSCOCO, with the number of prototypes $k$ set to 384 and 768, respectively. For experiments on ECCV Caption, we maintain the same hyperparameter settings as MSCOCO.

The method was implemented using PyTorch. Main experiments were conducted on a single NVIDIA RTX A6000 GPU, while retrieval speed tests used a single NVIDIA GeForce RTX 3090 GPU.
\begin{table*}[htbp]
	\centering
	\caption{
		Comparison with state-of-the-art methods on Flickr30K and MSCOCO 1k test sets. $\ast$ indicates the ensemble results of two models. The best results are shown in bold.}
	\label{tab:comparison1k}
	\resizebox{\textwidth}{!}{%
		\begin{tabular}{lc cccccc c cccccc c}
			\toprule[1pt]
			\multirow{3}{*}{Methods} & \multirow{3}{*}{Venue} & \multicolumn{7}{c}{Flickr30K dataset} & \multicolumn{7}{c}{MSCOCO (1k) dataset} \\
			\cline{3-16}
			\addlinespace[3pt]
			& & \multicolumn{3}{c}{IMG $\rightarrow$ TXT} & \multicolumn{3}{c}{TXT $\rightarrow$ IMG} & \multirow{2}{*}{rSum} & \multicolumn{3}{c}{IMG $\rightarrow$ TXT} & \multicolumn{3}{c}{TXT $\rightarrow$ IMG} & \multirow{2}{*}{rSum} \\
			& & R@1 & R@5 & R@10 & R@1 & R@5 & R@10 & & R@1 & R@5 & R@10 & R@1 & R@5 & R@10 & \\
			\hline
			\multicolumn{16}{l}{\textit{Interactive-embedding methods}} \\[0.5ex]
			SCAN* \cite{SCAN} & ECCV-18 & 67.4 & 90.3 & 95.8 & 48.6 & 77.7 & 85.2 & 465.0 & 72.7 & 94.8 & 98.4 & 58.8 & 88.4 & 94.8 & 507.9 \\ \addlinespace[1pt]
			IMRAM* \cite{IMRAM} & CVPR-20 & 74.1 & 93.0 & 96.6 & 53.9 & 79.4 & 87.2 & 484.2 & 76.7 & 95.6 & 98.5 & 61.7 & 89.1 & 95.0 & 516.6 \\ \addlinespace[1pt]
			GSMN* \cite{GSMN} & CVPR-20 & 76.4 & 94.3 & 97.3 & 57.4 & 82.3 & 89.0 & 496.8 & 78.4 & 96.4 & 98.6 & 63.3 & 90.1 & 95.7 & 522.5 \\ \addlinespace[1pt]
			SGRAF* \cite{SGRAF} & AAAI-21 & 77.8 & 94.1 & 97.4 & 58.5 & 83.0 & 88.8 & 499.6 & 79.6 & 96.2 & 98.5 & 63.2 & 90.7 & 96.1 & 524.3 \\ \addlinespace[1pt]
			UARDA* \cite{UARDA} & TMM-22 & 77.8 & 95.0 & 97.6 & 57.8 & 82.9 & 89.2 & 500.3 & 78.6 & 96.5 & 98.9 & 63.9 & 90.7 & 96.2 & 524.8 \\ \addlinespace[1pt]
			DREN* \cite{DREN} & TCSVT-22 & 78.8 & 95.0 & 97.5 & 58.8 & 83.8 & 89.5 & 503.4 & 78.5 & 96.2 & 99.0 & 62.0 & 89.7 & 96.2 & 521.6 \\ \addlinespace[1pt]
			NAAF* \cite{NAAF} & CVPR-22 & 81.9 & 96.1 & 98.3 & 61.0 & 85.3 & 90.6 & 513.2 & 78.1 & 96.1 & 98.6 & 63.5 & 89.6 & 95.3 & 521.2 \\ \addlinespace[1pt]
			RCAR \cite{RCAR} & TIP-23 & 77.8 & 93.6 & 96.9 & 57.2 & 82.8 & 88.5 & 496.8 & 78.2 & 96.3 & 98.4 & 62.2 & 89.6 & 95.3 & 520.0 \\ \addlinespace[1pt]
			CHAN \cite{CHAN} & CVPR-23 & 80.6 & 96.1 & 97.8 & 63.9 & 87.5 & 92.6 & 518.5 & 81.4 & 96.9 & 98.9 & 66.5 & 92.1 & 96.7 & 532.6 \\ \addlinespace[1pt]
			TVRN \cite{TVRN} & TMM-24 & 82.1 & 95.6 & 98.3 & 63.9 & 87.6 & 92.6 & 520.1 & 81.1 & 96.4 & 98.8 & 67.7 & 92.3 & 97.1 & 533.4 \\
			\addlinespace[1pt]
			BOOM \cite{BOOM} & TCSVT-24 & 84.3 & 96.5 & 98.5 & 65.2 & 88.6 & 93.1 & 526.2 & 82.3 & 97.2 & 99.1 & 67.7 & 92.6 & 96.8 & 535.7 \\ \addlinespace[1pt]
			\hline
			\multicolumn{16}{l}{\textit{Independent-embedding methods}} \\[0.5ex]
			VSE++ \cite{VSE++} & BMVC-18 & 52.9 & 80.5 & 87.2 & 39.6 & 70.1 & 79.5 & 409.8 & 64.6 & 90.0 & 95.7 & 52.0 & 84.3 & 92.0 & 478.8 \\ \addlinespace[1pt]
			DSRAN* \cite{DSRAN} & TCSVT-20 & 77.8 & 95.1 & 97.6 & 59.2 & 86.0 & 91.9 & 507.6 & 78.3 & 95.7 & 98.4 & 64.5 & 90.8 & 95.8 & 523.5 \\ \addlinespace[1pt]
			CAMERA* \cite{CAMERA} & MM-20 & 78.0 & 95.1 & 97.9 & 60.3 & 85.9 & 91.7 & 508.9 & 75.9 & 95.5 & 98.6 & 62.3 & 90.1 & 95.2 & 517.6 \\ \addlinespace[1pt]
			CLIP (ViT-B/32) \cite{CLIP} & ICML-21 & 63.2 & 90.8 & 95.5 & 63.6 & 86.6 & 92.0 & 491.8 & 49.8 & 83.8 & 92.8 & 55.5 & 83.1 & 91.1 & 456.1 \\ \addlinespace[1pt]
			DIME* \cite{DIME} & SIGIR-21 & 81.0 & 95.9 & 98.4 & 63.6 & 88.1 & 93.0 & 520.0 & 78.8 & 96.3 & 98.7 & 64.8 & 91.5 & 96.5 & 526.6 \\ \addlinespace[1pt]
			GPO \cite{GPO} & CVPR-21 & 81.7 & 95.4 & 97.6 & 61.4 & 85.9 & 91.5 & 513.5 & 79.7 & 96.4 & 98.9 & 64.8 & 91.4 & 96.3 & 527.5 \\ \addlinespace[1pt]
			AME \cite{AME} & AAAI-22 & 78.4 & 95.4 & 97.8 & 62.1 & 86.8 & 91.9 & 512.4 & 78.6 & 96.0 & 98.6 & 64.2 & 90.3 & 95.7 & 523.4 \\ \addlinespace[1pt]
			VSRN++ \cite{VSRN++} & TPAMI-22 & 79.2 & 94.6 & 97.5 & 60.6 & 85.6 & 91.4 & 508.9 & 77.9 & 96.0 & 98.5 & 64.1 & 90.4 & 96.1 & 523.6 \\ \addlinespace[1pt]
			ESA* \cite{ESA} & TCSVT-23 & 84.6 & 96.6 & 98.6 & 66.3 & 88.8 & 93.1 & 528.0 & 81.0 & 96.9 & 98.9 & 66.4 & 92.2 & 96.5 & 531.9 \\ \addlinespace[1pt]
			HREM* \cite{HREM} & CVPR-23 & 84.0 & 96.1 & 98.6 & 64.4 & 88.0 & 93.1 & 524.2 & 82.9 & 96.9 & 99.0 & 67.1 & 92.0 & 96.6 & 534.6 \\ \addlinespace[1pt]
			CORA \cite{CORA} & CVPR-24 & 83.7 & 96.6 & 98.3 & 62.3 & 87.1 & 92.6 & 520.6 & 82.4 & 96.8 & 98.8 & 66.2 & 91.9 & 96.6 & 532.7 \\ \addlinespace[1pt]
			MVAAP \cite{MVAAP} & KBS-24 & 84.2 & 96.2 & 98.6 & 64.0 & 87.7 & 92.4 & 523.2 & 83.9 & 97.1 & 99.1 & 67.5 & 91.8 & 96.1 & 535.5 \\ \addlinespace[1pt]
			IMEB \cite{IMEB} & TCSVT-24 & 84.2 & 96.7 & 98.4 & 64.0 & 88.0 & 92.8 & 524.1 & 82.4 & 96.9 & 99.0 & 66.7 & 91.9 & 96.6 & 533.5 \\ \addlinespace[1pt]
			USER \cite{USER} & TIP-24 & 86.3 & 97.6 & \textbf{99.4} & 69.5 & 91.0 & 94.4 & 538.1 & 83.7 & 96.7 & 99.0 & 67.8 & 91.2 & 95.8 & 534.2 \\ \addlinespace[1pt]
			DVSE \cite{DVSE} & TMM-25 & 82.1 & 96.4 & 98.3 & 63.4 & 87.6 & 92.3 & 520.1 & 80.6 & 96.8 & 98.7 & 65.8 & 91.3 & 96.1 & 529.2 \\ \addlinespace[1pt]
			\textbf{AAHR} & This paper & \textbf{89.0} & \textbf{98.7} & \textbf{99.4} & \textbf{72.9} & \textbf{92.4} & \textbf{95.7} & \textbf{548.1} & \textbf{84.3} & \textbf{97.8} & \textbf{99.2} & \textbf{69.6} & \textbf{92.9} & \textbf{97.1} & \textbf{540.8} \\
			\bottomrule[1pt]
		\end{tabular}
	}
\end{table*}
\begin{table}[htbp]
	\centering
	\caption{Comparison with state-of-the-art methods on MSCOCO 5k test set. $\ast$ indicates the ensemble results of two models. The best results are shown in bold.}
	\label{tab:comparison5k}
	
	\begin{tabular*}{\textwidth}{@{\extracolsep{\fill}}l *{7}{c}}
		\toprule[1pt]
		\multirow{3}{*}{Methods} & \multicolumn{7}{c}{MSCOCO 5k dataset} \\
		\cline{2-8}
		\addlinespace[3pt]
		& \multicolumn{3}{c}{IMG $\rightarrow$ TEXT} & \multicolumn{3}{c}{TEXT $\rightarrow$ IMG} & \multirow{2}{*}{rSum} \\
		& R@1 & R@5 & R@10 & R@1 & R@5 & R@10 & \\
		\midrule
		\multicolumn{8}{l}{\textit{Interactive-embedding methods}} \\\addlinespace[1pt]
		SCAN* \cite{SCAN} & 50.4 & 82.2 & 90.0 & 38.6 & 69.3 & 80.4 & 410.9 \\ \addlinespace[1pt]
		IMRAM* \cite{IMRAM} & 53.7 & 83.2 & 91.0 & 39.7 & 69.1 & 79.8 & 416.5 \\ \addlinespace[1pt]
		SGRAF* \cite{SGRAF} & 57.8 & 84.9 & 91.6 & 41.9 & 70.7 & 81.3 & 428.2 \\ \addlinespace[1pt]
		UARDA* \cite{UARDA} & 56.2 & 83.8 & 91.3 & 40.6 & 69.5 & 80.9 & 422.3 \\\addlinespace[1pt]
		NAAF* \cite{NAAF} & 58.9 & 85.2 & 92.0 & 42.5 & 70.9 & 81.4 & 430.9 \\\addlinespace[1pt]
		RCAR \cite{RCAR} & 59.1 & 84.8 & 91.8 & 42.8 & 71.5 & 81.9 & 431.9 \\\addlinespace[1pt]
		CHAN \cite{CHAN} & 59.8 & 87.2 & 93.3 & 44.9 & 74.5 & 84.2 & 443.9 \\\addlinespace[1pt]
		TVRN \cite{TVRN} & 61.1 & 86.3 & 92.5 & 45.5 & 75.0 & 84.8 & 445.2 \\\addlinespace[1pt]
		BOOM \cite{BOOM} & 63.1 & 87.7 & 93.4 & 45.2 & 75.5 & 85.7 & 450.6 \\\addlinespace[1pt]
		\midrule
		\multicolumn{8}{l}{\textit{Independent-embedding methods}} \\ \addlinespace[1pt]
		VSE++ \cite{VSE++} & 41.3 & 71.1 & 81.2 & 30.3 & 59.4 & 72.4 & 409.8 \\ \addlinespace[1pt]
		CAMERA* \cite{CAMERA} & 53.1 & 81.3 & 89.8 & 39.0 & 70.5 & 81.5 & 415.2 \\ \addlinespace[1pt]
		DSRAN* \cite{DSRAN} & 55.3 & 83.5 & 90.9 & 41.7 & 72.7 & 82.8 & 426.9 \\ \addlinespace[1pt]
		CLIP (ViT-B/32) \cite{CLIP} & 27.5 & 54.7 & 69.9 & 35.9 & 62.0 & 72.4 & 322.4 \\ \addlinespace[1pt]
		GPO \cite{GPO} & 58.3 & 85.3 & 92.3 & 42.4 & 72.7 & 83.2 & 434.3 \\ \addlinespace[1pt]
		DIME* \cite{DIME} & 59.3 & 85.4 & 91.9 & 43.1 & 73.0 & 83.1 & 435.8 \\ \addlinespace[1pt]
		VSRN++ \cite{VSRN++} & 54.7 & 82.9 & 90.9 & 42.0 & 72.2 & 82.7 & 425.4 \\ \addlinespace[1pt]
		AME \cite{AME} & 57.1 & 83.5 & 91.6 & 42.2 & 71.7 & 82.0 & 428.1 \\ \addlinespace[1pt]
		ESA* \cite{ESA} & 61.1 & 86.6 & 92.9 & 43.9 & 74.1 & 84.4 & 443.0 \\ \addlinespace[1pt]
		HREM* \cite{HREM} & 64.0 & 88.5 & 93.7 & 45.4 & 75.1 & 84.3 & 450.9 \\ \addlinespace[1pt]
		CORA \cite{CORA} & 62.4 & 86.8 & 92.6 & 44.2 & 73.6 & 83.9 & 443.6 \\ \addlinespace[1pt]
		MVAAP \cite{MVAAP} & 62.2 & 87.3 & 93.3 & 44.0 & 73.7 & 83.6 & 444.1 \\ \addlinespace[1pt]
		IMEB \cite{IMEB} & 62.8 & 87.8 & 93.5 & 44.9 & 74.6 & 84.0 & 447.6 \\ \addlinespace[1pt]
		USER \cite{USER} & \textbf{67.6} & 88.4 & 93.5 & 47.7 & 75.1 & 83.7 & 456.0 \\ \addlinespace[1pt]
		DVSE \cite{DVSE} & 60.5 & 86.3 & 92.9 & 43.8 & 73.8 & 83.2 & 440.5 \\ \addlinespace[1pt]
		\textbf{AAHR} & 66.6 & \textbf{89.8} & \textbf{94.7} & \textbf{49.0} & \textbf{77.3} & \textbf{86.2} & \textbf{463.4} \\
		\bottomrule[1pt]
	\end{tabular*}%
	
\end{table}
\begin{table}[htbp]
	\centering
	\caption{Comparison with state-of-the-art methods on ECCV Caption test set. $\ast$ indicates the ensemble results of two models. The best results are shown in bold.}
	\label{tab:comparisonECCV}
	\resizebox{\columnwidth}{!}{%
		\begin{tabular*}{\textwidth}{@{\extracolsep{\fill}}l *{7}{c}}
			\toprule[1pt]
			\multirow{3}{*}{Methods} & \multicolumn{7}{c}{ECCV Caption dataset} \\
			\cline{2-8}
			\addlinespace[3pt]
			& \multicolumn{3}{c}{IMG $\rightarrow$ TEXT} & \multicolumn{3}{c}{TEXT $\rightarrow$ IMG} & \multirow{2}{*}{Sum} \\
			& mAP@R & R-P & R@1 & mAP@R & R-P & R@1 & \\
			\midrule
			CLIP (ViT-B/32) \cite{CLIP} & 22.4 & 32.6 & 66.1 & 31.1 & 41.2 & 68.1 & 261.5 \\ \addlinespace[1pt]
			CLIP (ViT-L/14) \cite{CLIP} & 24.0 & 33.8 & 71.4 & 32.0 & 41.8 & 73.0 & 275.9 \\ \addlinespace[1pt]
			SGR \cite{SGRAF} & 27.2 & 39.2 & 71.1 & 44.4 & 52.9 & 86.5 & 321.2 \\ \addlinespace[1pt]
			SAF \cite{SGRAF} & 27.4 & 39.3 & 71.1 & 44.6 & 53.1 & 85.7 & 321.0 \\ \addlinespace[1pt]
			SGRAF* \cite{SGRAF} & 28.6 & 40.3 & 73.8 & 46.0 & 54.3 & 87.8 & 330.8 \\ \addlinespace[1pt]
			GPO \cite{GPO} & 31.4 & 42.8 & 74.9 & 49.6 & 57.2 & 90.2 & 345.9 \\ \addlinespace[1pt]
			CHAN \cite{CHAN} & 30.5 & 41.9 & 75.2 & 46.4 & 54.3 & 88.2 & 336.5 \\ \addlinespace[1pt]
			HREM \cite{HREM} & 31.5 & 42.5 & 78.4 & 47.9 & 55.9 & 88.1 & 344.3 \\ \addlinespace[1pt]
			ESL \cite{ESL} & 30.9 & 42.0 & 77.3 & 45.7 & 53.7 & 88.1 & 337.6 \\ \addlinespace[1pt]
			ESA \cite{ESA} & 31.5 & 43.0 & 74.6 & 49.3 & 56.9 & 90.2 & 345.5 \\ \addlinespace[1pt]
			AAHR$_{\text{w/o PGA}}$ & 33.7 & 44.6 & 80.7 & 49.3 & 57.0 & 89.0 & 354.3 \\ \addlinespace[1pt]
			AAHR & \textbf{34.2} & \textbf{44.8} & \textbf{81.3} & \textbf{49.7} & \textbf{57.4} & \textbf{90.3} & \textbf{357.7} \\
			\bottomrule[1pt]
		\end{tabular*}%
	}
\end{table}
\subsection{Comparison with State-of-the-art Methods}
To validate the effectiveness of our proposed method, we conducted a comprehensive comparison with state-of-the-art approaches on several benchmark datasets. These methods can be broadly categorized into local matching methods based on interactive embeddings and global matching methods based on independent embeddings. Tables \ref{tab:comparison1k}, \ref{tab:comparison5k} and \ref{tab:comparisonECCV} present detailed quantitative results. Some ensemble methods (marked with $\ast$) improve performance by averaging the similarities of different models.

Our method comprehensively outperforms existing state-of-the-art approaches on the key metrics of R@K, mAP@R, R-P and rSum, except for the R@1 metric for image-to-text retrieval in Table \ref{tab:comparison5k}. These results demonstrate our proposed method's superior performance and broad applicability.

Our method demonstrates exceptional performance on the Flickr30K dataset. Compared to the current state-of-the-art interactive embedding method, BOOM, and the state-of-the-art independent embedding method, USER, we achieve relative improvements of 21.9\% and 10\% on the rSum metric, respectively. Notably, when compared to our strongest competitor, USER, our method consistently outperforms in R@K (K=1,5,10) metrics for both text retrieval and image retrieval tasks. Specifically, for image-to-text retrieval, we show improvements of 2.7\%, 1.1\%, and 0.05\%, respectively; for text-to-image retrieval, the improvements are 3.4\%, 1.4\%, and 1.3\%, respectively.

On MSCOCO, our approach exhibits outstanding performance on both 1K and 5K test sets, with more significant advantages on the larger and more challenging 5K dataset. These results validate our method's effectiveness, robustness, and generalization capability for large-scale datasets.

On ECCV Caption dataset, our method AAHR achieves new state-of-the-art performance across all evaluation metrics. In image-to-text retrieval, our approach achieves mAP@R of 34.2\%, R-P of 44.8\%, and R@1 of 81.3\%, surpassing the previous best results by 2.7\%, 1.8\%, and 2.9\%, respectively. In text-to-image retrieval, we attain mAP@R of 49.7\%, R-P of 57.4\%, and R@1 of 90.3\%. The overall Sum metric reaches 357.7. Notably, the comparison between AAHR and AAHR$_{\text{w/o PGA}}$ highlights the substantial contribution of the PGA module in addressing the challenge of soft positive sample retrieval. By generating representative category prototypes as shared semantic anchors, the PGA module effectively enhances the model's capability to identify semantically similar instances. These results are particularly meaningful on ECCV Caption, given its expanded positive pairs and comprehensive evaluation metrics, which better reflect real-world retrieval scenarios. The consistent improvements across both traditional R@K and the more rigorous mAP@R and R-P metrics demonstrate our method's robust capability in handling complex matching relationships.

Significant performance improvements are achieved by our method on several public image-text matching benchmarks. Our performance improvements do not solely stem from CLIP model knowledge transfer. We use a frozen CLIP (ViT/B-32) model without fine-tuning, which alone does not constitute a strong baseline (as shown in Tables \ref{tab:comparison1k}, \ref{tab:comparison5k} and \ref{tab:comparisonECCV}). We consistently outperform USER, which also employs CLIP, across almost all evaluation metrics.

Our significant progress can be attributed to the synergistic effect among various modules, which will be demonstrated explicitly in the ablation studies section \ref{Ablation}.
\begin{table*}[thbp]
	\centering
	\caption{The ablation study on framework design was conducted on Flickr30K. In the table, “$\checkmark$” indicates that the component is retained; otherwise, it is removed.}
	\label{tab:ablation}
	\resizebox{\textwidth}{!}{%
		\begin{tabular}{ccccccccccccccc}
			\toprule[1pt]
			\multirow{2}{*}{Methods} & \multirow{2}{*}{Local} & \multirow{2}{*}{Global} & \multirow{2}{*}{GGLA} & \multirow{2}{*}{MDM} & \multirow{2}{*}{PGA} & \multirow{2}{*}{NSI} & \multirow{2}{*}{rSum} & \multicolumn{3}{c}{IMG $\rightarrow$ TEXT} & \multicolumn{3}{c}{TEXT $\rightarrow$ IMG} \\
			\cmidrule(lr){9-11} \cmidrule(l){12-14}
			&  &  &  &  &  &  &  & R@1 & R@5 & R@10 & R@1 & R@5 & R@10 \\
			\midrule
			1 & \checkmark &  &  &  &  &  & 515.8 & 81.6 & 95.6 & 98.2 & 62.3 & 86.3 & 91.8 \\ \addlinespace[1pt]
			2 & \checkmark &  &  & \checkmark &  &  & 517.2 & 81.4 & 96.1 & 98.0 & 63.0 & 86.7 & 92.0 \\ \addlinespace[1pt]
			3 & \checkmark &  &  & \checkmark & \checkmark &  & 517.3 & 82.3 & 96.4 & 98.4 & 62.1 & 86.2 & 91.8 \\ \addlinespace[1pt]
			4 & \checkmark &  &  & \checkmark & \checkmark & \checkmark & 518.8 & 82.4 & 96.4 & 97.8 & 63.1 & 86.8 & 92.4 \\ \addlinespace[1pt]
			5 &  & \checkmark &  &  &  &  & 532.7 & 84.3 & 96.6 & 98.2 & 69.7 & 90.0 & 93.9 \\ \addlinespace[1pt]
			6 &  & \checkmark &  & \checkmark &  &  & 533.2 & 85.0 & 96.6 & 98.2 & 69.7 & 89.7 & 94.0 \\ \addlinespace[1pt]
			7 &  & \checkmark &  & \checkmark & \checkmark &  & 533.6 & 84.8 & 96.6 & 98.3 & 70.1 & 89.9 & 94.0 \\ \addlinespace[1pt]
			8 &  & \checkmark &  & \checkmark & \checkmark & \checkmark & 534.1 & 85.2 & 96.7 & 98.2 & 69.7 & 90.3 & 94.0 \\ \addlinespace[1pt]
			9 & \checkmark & \checkmark &  & \checkmark & \checkmark & \checkmark & 547.1 & 88.1 & 98.5 & \textbf{99.6} & \textbf{73.1} & \textbf{92.4} & 95.4 \\ \addlinespace[1pt]
			10 & \checkmark & \checkmark & \checkmark & \checkmark & \checkmark & \checkmark & \textbf{548.1} & \textbf{89.0} & \textbf{98.7} & 99.4 & 72.9 & \textbf{92.4} & \textbf{95.7} \\
			\bottomrule[1pt]
		\end{tabular}%
	}
\end{table*}
\begin{table}[htbp]
	\centering
	\caption{Performance comparison of different dynamic memory bank sizes on MSCOCO 1K and MSCOCO 5K test sets.}
	\label{tab:MDM_size}
	\begin{tabular*}{\textwidth}{@{\extracolsep{\fill}}c *{7}{c}}
		\toprule[1pt]
		\multirow{2}{*}{Size} & \multicolumn{3}{c}{IMG $\rightarrow$ TEXT} & \multicolumn{3}{c}{TEXT $\rightarrow$ IMG} & \multirow{2}{*}{rSum} \\
		\cmidrule(lr){2-7}
		& R@1 & R@5 & R@10 & R@1 & R@5 & R@10 & \\
		\midrule
		\multicolumn{8}{l}{\textit{MSCOCO 1K Testing set}} \\[0.5ex]
		256 & 82.8 & 97.1 & 99.1 & 67.3 & 91.8 & 96.6 & 534.7 \\ \addlinespace[1pt]
		1024 & 84.0 & 97.6 & 99.1 & 68.9 & 92.6 & 96.9 & 539.2 \\ \addlinespace[1pt]
		\textbf{2048} & \textbf{84.3} & \textbf{97.8} & \textbf{99.2} & \textbf{69.6} & \textbf{92.9} & \textbf{97.1} & \textbf{540.8} \\ \addlinespace[1pt]
		4096 & 83.5 & 97.3 & 99.2 & 68.0 & 92.2 & 96.8 & 537.1 \\ \addlinespace[1pt]
		\midrule
		\multicolumn{8}{l}{\textit{MSCOCO 5K Testing set}} \\[0.5ex]
		256 & 64.8 & 87.5 & 93.5 & 46.1 & 74.9 & 84.2 & 451.1 \\ \addlinespace[1pt]
		1024 & \textbf{66.7} & 89.6 & \textbf{94.7} & 48.1 & 76.8 & 85.6 & 461.4 \\ \addlinespace[1pt]
		\textbf{2048} & 66.6 & \textbf{89.8} & \textbf{94.7} & \textbf{49.0} & \textbf{77.3} & \textbf{86.2} & \textbf{463.4} \\ \addlinespace[1pt]
		4096 & 66.0 & 88.9 & 94.3 & 46.9 & 75.6 & 84.6 & 456.2 \\
		\bottomrule[1pt]
	\end{tabular*}
\end{table}
\subsection{Ablation Studies}\label{Ablation}
To validate the effectiveness of each component in our proposed method, we conducted comprehensive ablation experiments. Table \ref{tab:ablation} presents the results, where Local and Global refer to different feature extraction approaches, GGLA represents the Global semantics-Guided adaptive Local feature Aggregation module, MDM denotes the Momentum Dynamic Memory bank, PGA stands for the Prototype-Guided cross-modal Alignment module, and NSI represents the Neighborhood Semantic Interaction module. The “$\checkmark$” in the table indicates that the component is retained ;otherwise, it is removed.
\subsubsection{Complementarity of Local and Global features}
Our analysis first focused on the impact of feature extraction strategies. Table \ref{tab:ablation} shows its performance is relatively weak when the model uses only local or global features. We explored the complementarity of local and global features, a crucial finding in this study. While the classical work TERAN \cite{TERAN} suggests that the integration of global alignment with local alignment may trigger optimization conflicts, our study finds significant complementarity between local and global features from a different perspective. The introduction of the GGLA module, which integrates these complementary features, significantly improved model performance. Specifically, the rSum metric of Method 10 improved by 29.3\% and 14\% compared to the best methods using only local features (Method 4) and only global features (Method 8), respectively.

\subsubsection{Effectiveness of the GGLA}
Further analysis reveals that Method 10 achieved comprehensive improvements compared to Method 9, in which local features are aggregated using mean pooling. This powerfully demonstrates the GGLA module's superior performance in feature fusion. Local features excel at capturing fine-grained regional information, while global features have advantages in understanding overall semantics. The GGLA module intelligently integrates these two features, providing the model with more comprehensive and powerful representation capabilities, significantly enhancing cross-modal retrieval task performance.
\subsubsection{Effectiveness of the PGA}
Comparisons between Methods 2 and 3, and 6 and 7 show significant improvements in most metrics with the introduction of the PGA module. PGA generates representative category prototypes, providing common reference points for features from different modalities. This reduces semantic discrepancies between modalities, enabling more consistent and robust cross-modal representations. PGA's effect is more pronounced on global features, indicating better synergy between cluster prototypes and global semantic information. As shown in Table \ref{tab:comparisonECCV}, the introduction of PGA significantly enhances performance across all metrics, achieving mAP@R of 34.2\% and 49.7\% for image-to-text and text-to-image retrieval, respectively. This substantial improvement demonstrates the effectiveness of PGA's prototype-based alignment strategy in addressing the soft positive sample challenge.
\subsubsection{Effectiveness of the NSI}
Comparisons of Methods 3 with 4 and 7 with 8 demonstrate performance improvements brought by the NSI module. NSI enhances feature representation by leveraging semantic relationships between instances, capturing intricate semantic structures through neighborhood information in the feature space. The significant improvement in the R@10 metric highlights NSI's advantage in enhancing retrieval comprehensiveness.
\subsubsection{Effectiveness of the MDM}
Comparing Methods 1 with 2 and 5 with 6 reveals consistent performance improvements when using the Momentum Dynamic Memory bank (MDM). As a dynamic learning repository, MDM stores numerous image and text feature representations as negative samples for contrastive learning, enhancing feature discriminative ability. Unlike static structures, MDM continuously updates stored information throughout training.
\subsubsection{Momentum Dynamic Memory Bank Size}
To evaluate the impact of the Momentum Dynamic Memory (MDM) bank size on retrieval performance, we conducted experiments on both MSCOCO 5K and MSCOCO 1K test sets.
The results in Table \ref{tab:MDM_size} demonstrate that the MDM size significantly influences the retrieval performance. As the MDM size increases from 256 to 2,048, the rSum metric shows a consistent upward trend for both test sets, with 6.1\% and 12.3\% improvements for the 1K and 5K test sets, respectively. This suggests that a larger memory bank can accommodate a more diverse range of negative samples, helping the model learn more robust feature representations. However, increasing MDM size to 4,096 results in a slight performance decline, possibly due to excessive noise. The impact of MDM size variations is more pronounced on the 5K test set, indicating advantages of larger MDM when handling larger or more complex datasets.
\begin{figure}[tbp]
	\centering
	\subfigure[Number of cluster prototypes $k$]{
		\begin{minipage}[t]{0.48\linewidth}
			\centering
			\includegraphics[width=1\linewidth]{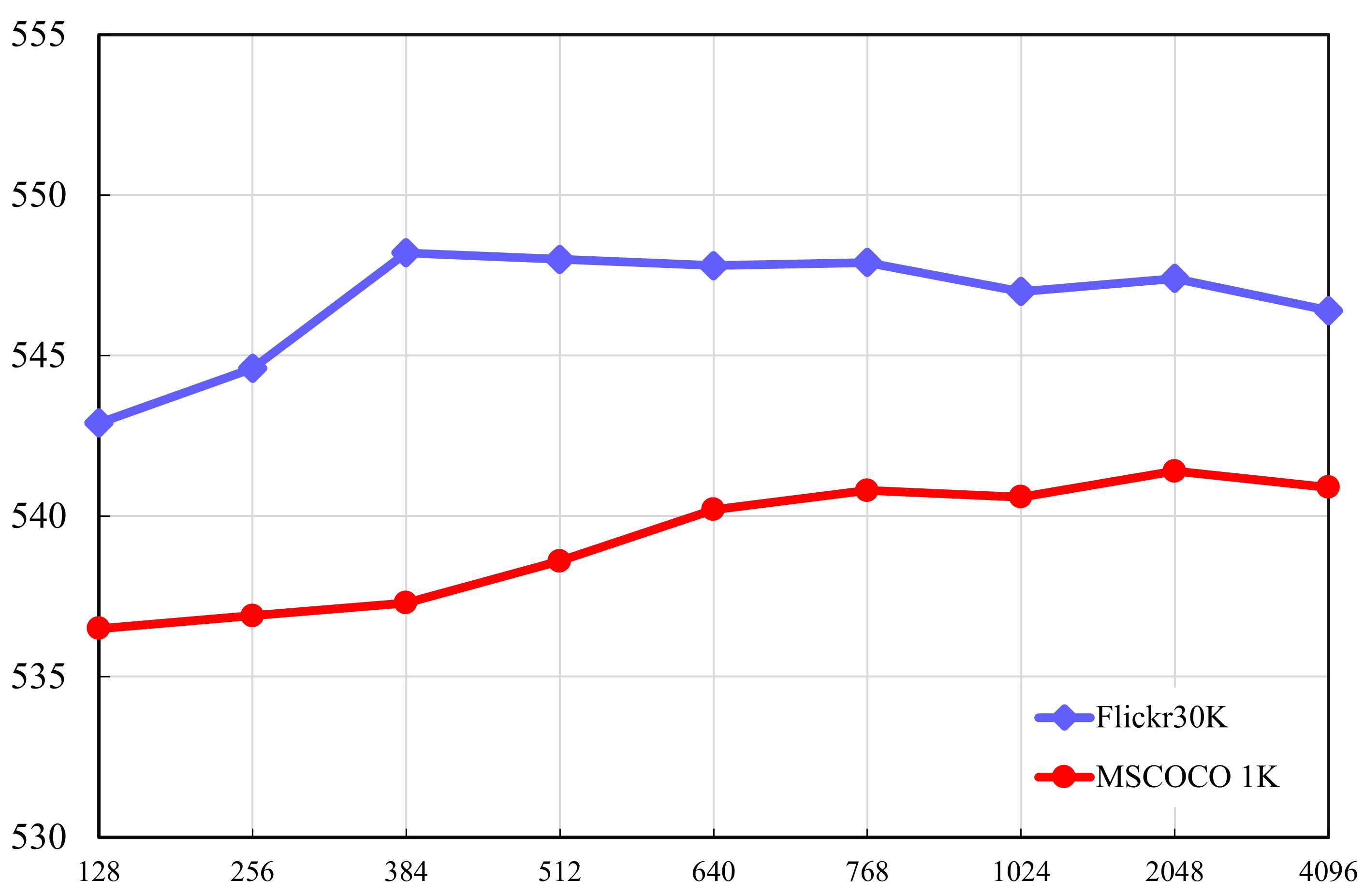}
		\end{minipage}%
	}%
	\subfigure[Amplification coefficient $\alpha$]{
		\begin{minipage}[t]{0.48\linewidth}
			\centering
			\includegraphics[width=1\linewidth]{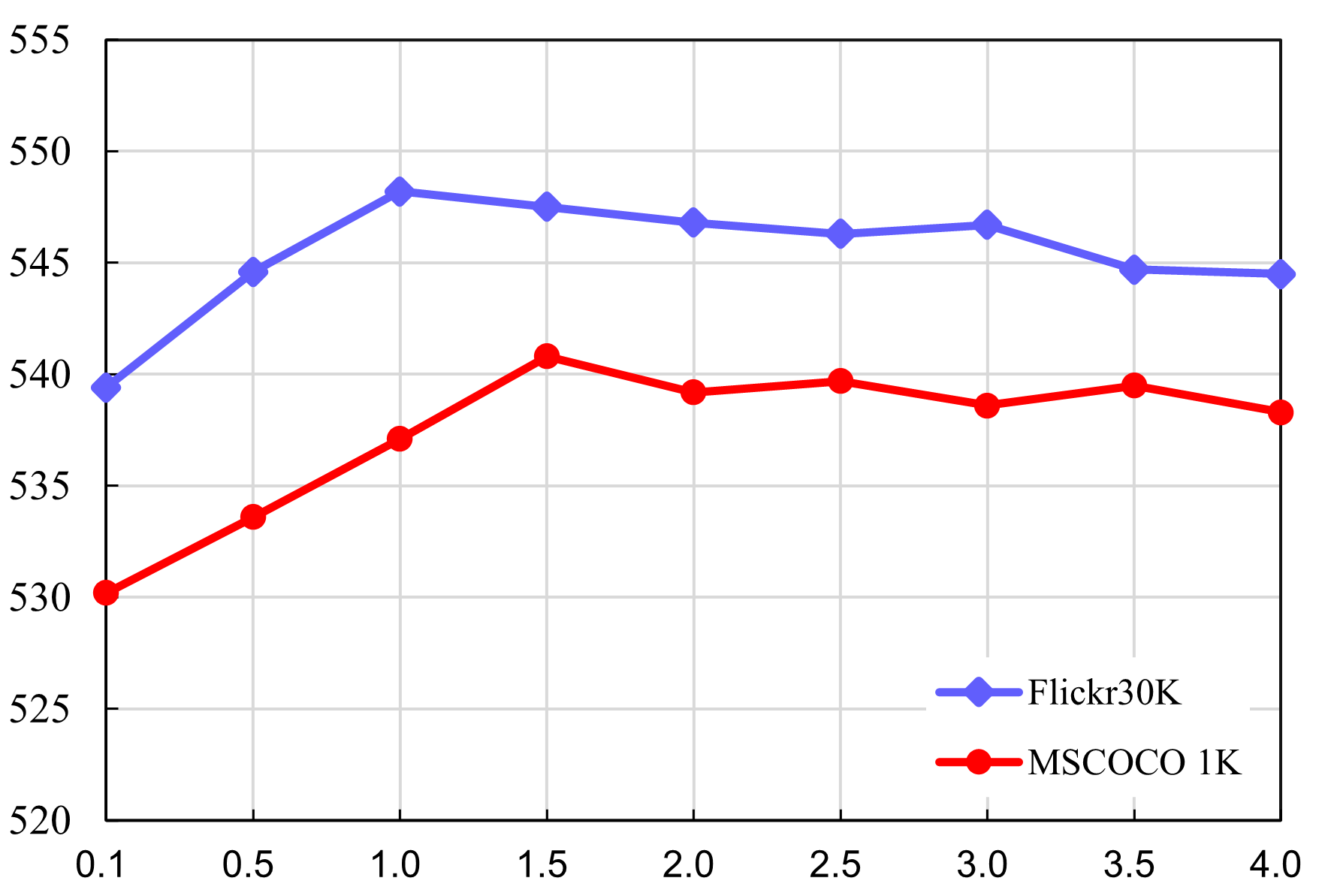}
		\end{minipage}%
	}%
	\centering
	\caption{Parameter analyses of $k$ in Eq.(\ref{PGA_img}, \ref{PGA_txt}) and $\alpha$ in Eq.(\ref{at}) on both datasets. The y-axis represents rSum (Recall Sum) for both plots.}
	\label{fig:Param_analysis}
\end{figure}
\begin{table}[htbp]
	\centering
	\caption{Comparison of model performance trained on MSCOCO and evaluated on the Flickr30K test set.}
	\label{tab:corss_dataset}
	\small
	\begin{tabular*}{\textwidth}{@{\extracolsep{\fill}}l *{7}{c}}
		\toprule[1pt]
		\multirow{2}{*}{Methods} & \multicolumn{3}{c}{IMG $\rightarrow$ TEXT} & \multicolumn{3}{c}{TEXT $\rightarrow$ IMG} & \multirow{2}{*}{rSum} \\
		\cmidrule(lr){2-7}
		& R@1 & R@5 & R@10 & R@1 & R@5 & R@10 & \\
		\midrule[0.5pt]
		SCAN \cite{SCAN} & 49.8 & 77.8 & 86.0 & 38.4 & 65.0 & 74.4 & 391.4 \\ \addlinespace[1pt]
		SGRAF \cite{SGRAF} & 65.7 & 87.2 & 93.4 & 48.1 & 73.9 & 81.9 & 450.2 \\ \addlinespace[1pt]
		GPO \cite{GPO} & 68.0 & 89.2 & 93.7 & 50.0 & 77.0 & 84.9 & 462.8 \\ \addlinespace[1pt]
		ESA \cite{ESA} & 69.5 & 89.1 & 93.8 & 51.5 & 77.9 & 85.7 & 467.4 \\ \addlinespace[1pt]
		HREM \cite{HREM} & 70.3 & 90.7 & 95.2 & 52.9 & 78.7 & 86.2 & 474.0\\ \addlinespace[1pt]
		CHAN \cite{CHAN} & 68.7 & 91.5 & 95.3 & 55.6 & 81.2 & 87.5 & 479.8 \\ \addlinespace[1pt]
		IMEB \cite{IMEB} & 72.3 & 90.2 & 95.1 & 54.8 & 80.0 & 86.9 & 479.3 \\ \addlinespace[1pt]
		ESL \cite{ESL} & 73.1 & 93.5 & 95.6 & 55.6 & 80.3 & 87.1 & 485.3 \\ \addlinespace[1pt]
		BOOM \cite{BOOM} & 72.0 & 91.9 & 96.0 & 57.4 & 82.2 & 88.6 & 488.1 \\ \addlinespace[1pt]
		\textbf{AAHR} & \textbf{85.5} & \textbf{98.4} & \textbf{99.4} & \textbf{71.5} & \textbf{93.7} & \textbf{97.2} & \textbf{545.6} \\
		\bottomrule[1pt]
	\end{tabular*}
\end{table}
\subsubsection{Hyperparameters Analysis}
Fig. \ref{fig:Param_analysis} illustrates the impact of two key parameters on retrieval performance, where $k$ represents the number of cluster prototypes, and $\alpha$ is the semantic amplification coefficient for the multimodal instance neighborhood semantic association graph. It can be observed that the performance of the model remains relatively stable across these parameter variations, demonstrating robust hyperparameter insensitivity.
\subsection{Further Analysis}
\subsubsection{Generalization Capability}
We conducted zero-shot transfer experiments across datasets to further validate the model's generalization ability. Specifically, we tested the performance of models trained on the MSCOCO when applied to the Flickr30K. As shown in Table \ref{tab:corss_dataset}, our method demonstrates excellent performance in this task, proving that our model can effectively capture universal semantic relationships across datasets and highlighting the superiority of our proposed approach.
\begin{figure*}[t]
	\centering
	\includegraphics[width=\textwidth]{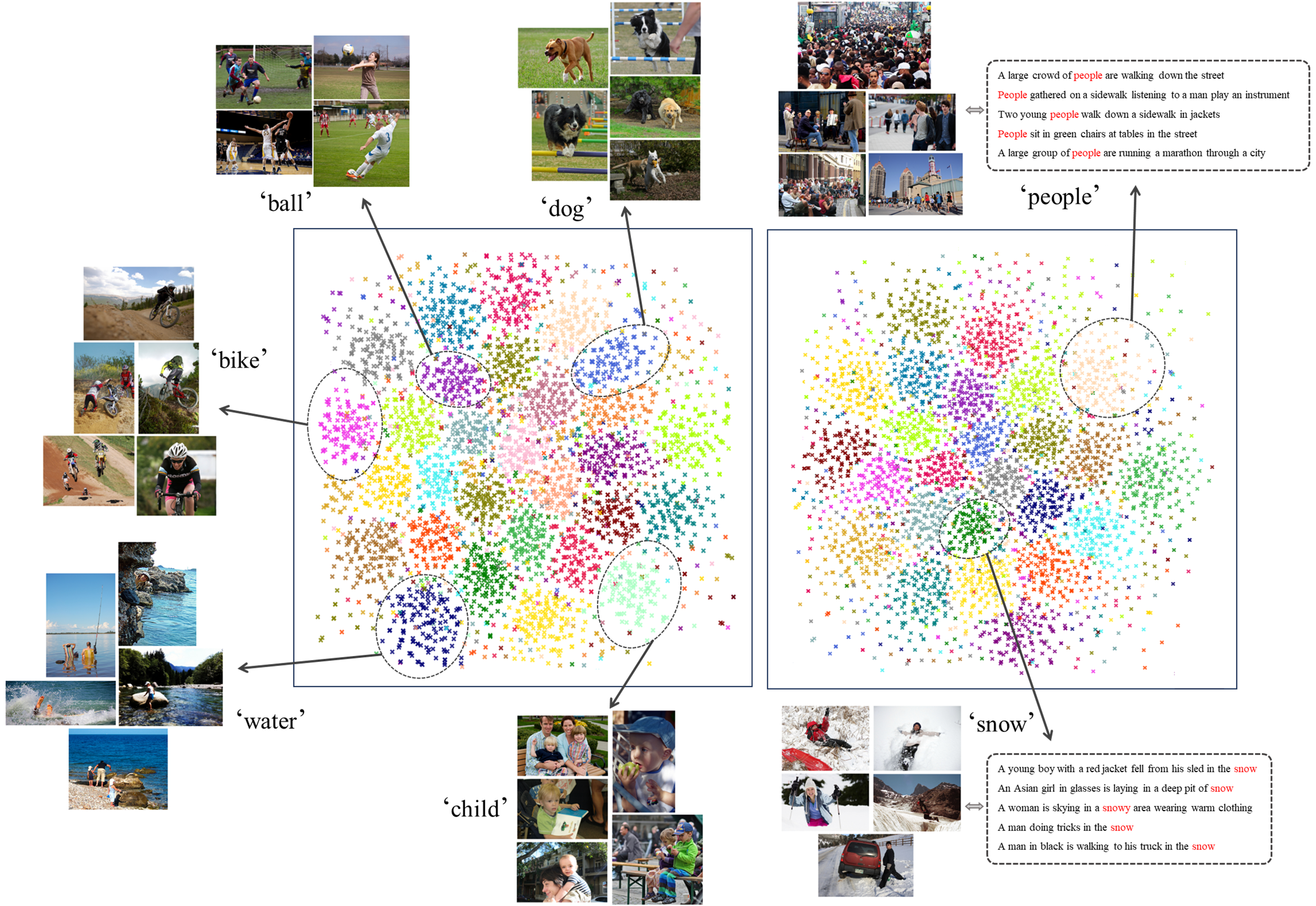}
	\caption{Dimensionality reduction visualization of feature embeddings using the t-SNE algorithm. The left figure shows the distribution of visual embeddings, while the right figure displays the distribution of textual embeddings. Clusters of points with the same color represent representations with identical or similar semantic concepts.
	} \label{fig:tsne}
\end{figure*}
\begin{figure}[htbp]
	\centering
	\includegraphics[width=\linewidth]{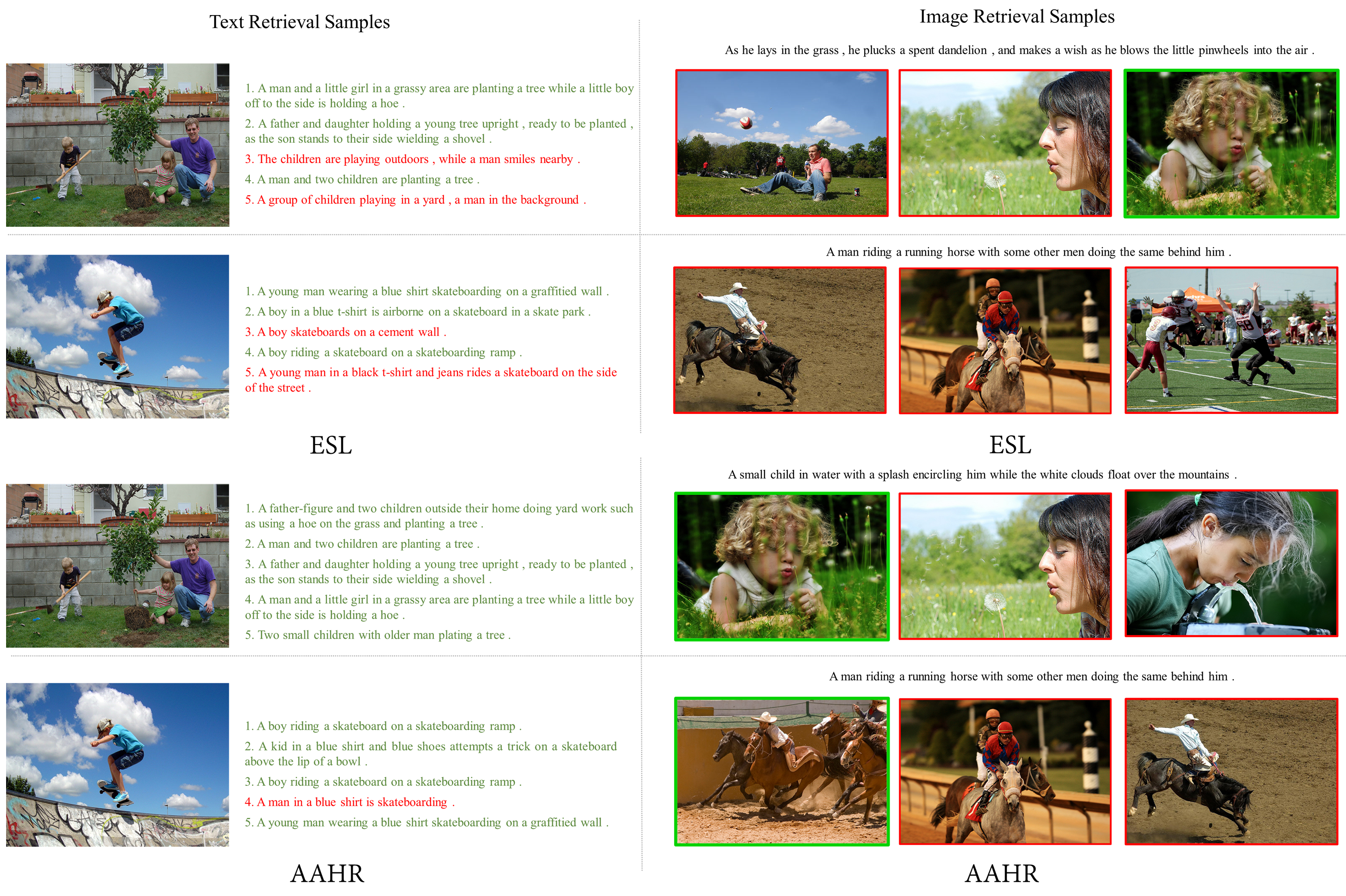}
	\caption{This figure presents the results of bidirectional image-text retrieval experiments conducted on the Flickr30K dataset. Green text or borders indicate results that align with the dataset's ground truth, while red suggests inconsistencies.
	} \label{fig:retrieve_case_study}
\end{figure}
\begin{figure}[tbp]
	\centering
	\includegraphics[width=\linewidth]{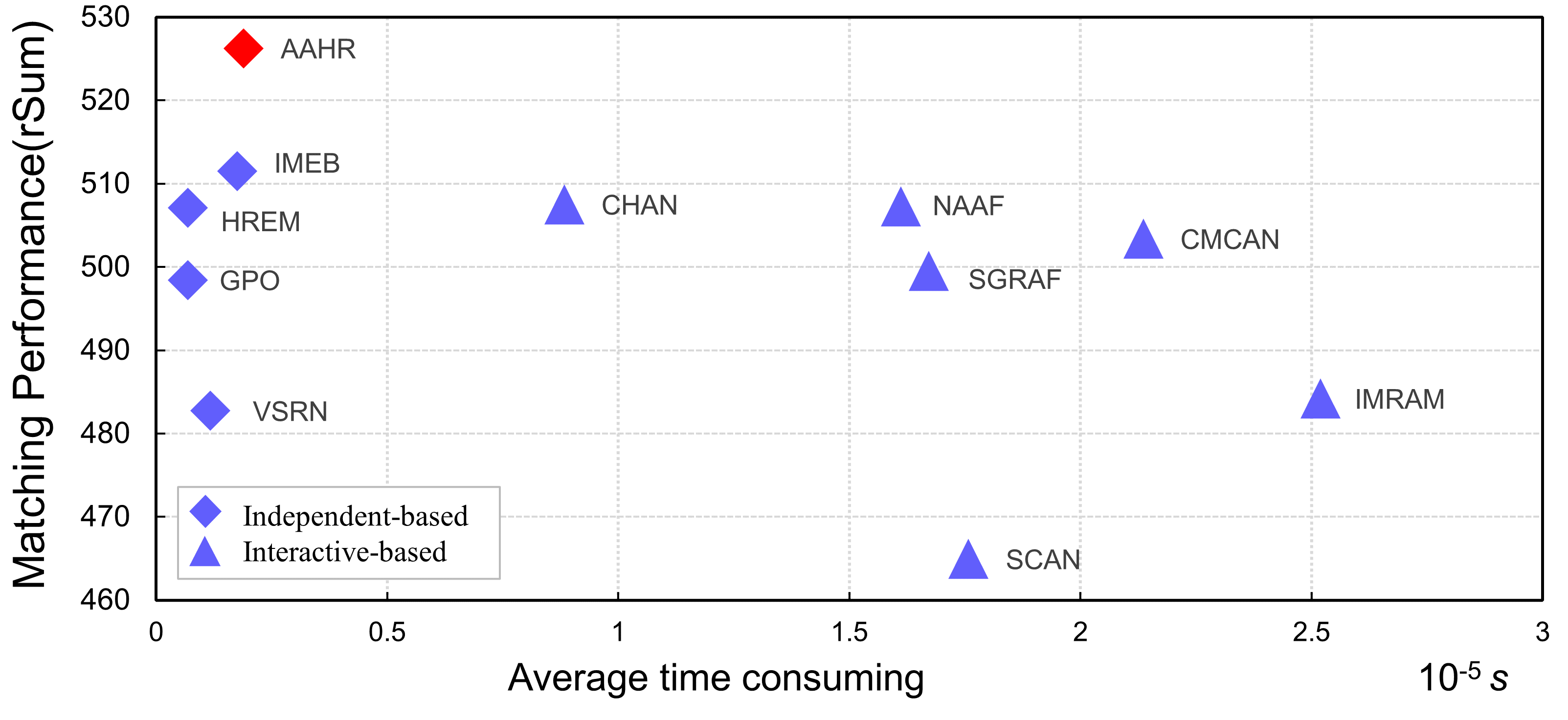}
	\caption{Comparison of accuracy and speed. All methods were executed on the Flickr30K test set using a single RTX3090 GPU.
	} \label{fig:Retrieval_efficiency}
\end{figure}
\begin{figure*}[t]
	\centering
	\includegraphics[width=\textwidth]{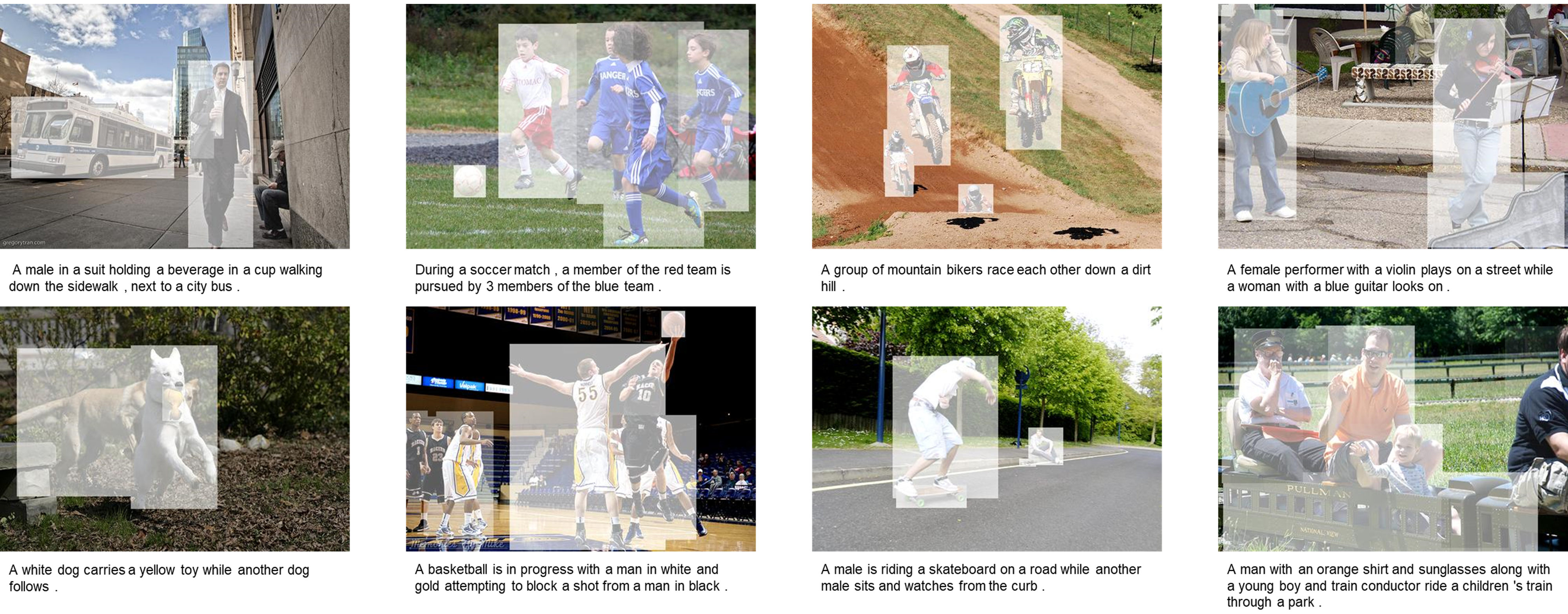}
	\caption{Visualization of semantic-focused attention on salient image regions. The transparency is inversely proportional to the attention intensity, with low-transparency areas indicating high-attention regions. The results demonstrate that our method effectively captures core semantic elements.
	} \label{fig:att_visual}
\end{figure*}%
\subsubsection{Retrieval Efficiency} Fig. \ref{fig:Retrieval_efficiency} compares performance and computation time for cross-modal retrieval on the Flickr30K test set using an RTX3090 GPU. Our method achieves an excellent balance between accuracy and efficiency, significantly outperforming other methods in performance while maintaining extremely low average retrieval times.
\subsection{Visualization and Case Analysis}
\subsubsection{Visualization of Embedding Distributions}
To further validate the rationality and effectiveness of our proposed method, we employed the t-SNE algorithm \cite{tSNE} to visualize the dimensionality reduction of the extracted multimodal feature embeddings. Fig. \ref{fig:tsne} illustrates the distribution of visual and textual embeddings learned by our method in two-dimensional space based on the Flickr30K test set. The visualization shows distinct boundaries between different semantic categories with low overlap between clusters, demonstrating that our method can learn highly discriminative feature representations.
\subsubsection{Visualization of Retrieval Results}
To intuitively validate the generalization capability of our method, we conducted cross-dataset evaluation by directly transferring the model trained on MSCOCO dataset to Flickr30K test set for zero-shot testing. As illustrated in the bidirectional retrieval visualization comparison in Fig. \ref{fig:retrieve_case_study}, our proposed AAHR method demonstrates significant performance improvements over the baseline ESL model. Notably, in the image-to-text retrieval task, both methods retrieved some captions labeled as "incorrect" but semantically relevant (i.e., soft negative samples). This phenomenon reveals inherent biases in the dataset annotation.%
\subsubsection{Visualization of Semantic Focus Attention}
To evaluate the ability of our full-granularity feature aggregation method to capture core scene semantics, we designed a semantic focus attention visualization experiment. We generated visualizations highlighting high-attention areas by calculating the attention scores between the image's full-granularity aggregated features and salient region features. As shown in Fig. \ref{fig:att_visual},  our method precisely locates key semantic elements in images, aligning with human visual cognition patterns. This visualization verifies the method's effectiveness in understanding complex visual scenes, reflecting the synergistic effect of global and local semantic features.
\section{Conclusion}
This research focuses on a frequently overlooked core issue in image-text matching: higher-order associations and semantic ambiguities among similar instances. Existing methods often mislabel semantically similar instances as negative samples during data annotation, forming “soft positive samples”, which introduce additional noise and confusion in model training. Furthermore, during the inference stage, models tend to misidentify instances with good local alignment but mismatched overall semantics as positive samples, \textit{i.e.}, “soft negative samples”, significantly reducing matching precision. Concurrently, training batches contain numerous semantically similar instances, embodying rich higher-order shared knowledge. We propose the Ambiguity-Aware and Higher-order Relation learning framework (AAHR) to address these challenges. This framework effectively captures higher-order associations and subtle differences between instances at different stages, improving matching accuracy. Extensive experimental results confirm the significant advantages of our method. In the future, we plan to extend this approach to broader multimodal tasks, such as video-text retrieval and unified multimodal retrieval, to further validate its applicability and effectiveness in complex scenarios.
\bibliographystyle{unsrt}
\bibliography{cas-refsv1}

\end{document}